\RequirePackage{fix-cm}
\documentclass[smallcondensed]{svjour3}       
\smartqed  
\usepackage{graphicx}
\usepackage{color,soul}
\usepackage{lipsum}  
\usepackage{graphicx}
\usepackage[utf8]{inputenc}
\usepackage{amsmath}
\usepackage{amsfonts}
\usepackage{amssymb}
\usepackage{tikz}
\usetikzlibrary{fit,positioning,quotes,arrows.meta}
\usepackage{algorithm}
\usepackage{booktabs}
\usepackage{mathtools}
\usepackage{algpseudocode}
\usepackage{blindtext}
\usepackage{xcolor}
\usepackage{bm}
\usepackage{tikz}
\usetikzlibrary{bayesnet}
\usetikzlibrary{arrows}
\DeclareMathOperator*{\argmin}{arg\,min}
\DeclareMathOperator*{\argmax}{arg\,max}
\DeclareMathOperator{\E}{\mathbb{E}}
\DeclareMathOperator{\U}{\mathbf{U}}
\DeclareMathOperator{\DKL}{\text{D}_\text{KL}}
\DeclareMathOperator*{\F}{\Delta \text{F}_{\text{par}}}

\DeclareMathAlphabet{\altmathcal}{OMS}{cmsy}{m}{n}
\usepackage{mathptmx}      
\journalname{Neural Processing Letters}

\begin{document}

\title{Specialization in Hierarchical Learning Systems}
\subtitle{A Unified Information-theoretic Approach for Supervised, Unsupervised and Reinforcement Learning}


\author{Heinke Hihn         \and
        Daniel A.\ Braun
}


\institute{Heinke Hihn  \at
              Institute for Neural Information Processing \\
              Ulm University, Ulm, Germany \\
              \email{heinke.hihn@uni-ulm.de}           
           \and
           Daniel  A.\ Braun \at
              Institute for Neural Information Processing \\
              Ulm University, Ulm, Germany\\
              \email{daniel.braun@uni-ulm.de}
}

\date{Received: date / Accepted: date}

\maketitle

\begin{abstract}
Joining multiple decision-makers  together is a powerful way to obtain more sophisticated decision-making systems, but requires to address the questions of division of labor and specialization.
We investigate in how far information constraints in hierarchies of experts not only provide a principled method for regularization but also to enforce specialization.
In particular, we devise an information-theoretically motivated on-line learning rule that allows partitioning of the problem space into multiple sub-problems that can be solved by the individual experts.  We demonstrate two different ways to apply our method: (i) partitioning problems based on individual data samples and (ii) based on sets of data samples representing tasks. Approach (i) equips the system with the ability to solve complex decision-making problems by finding an optimal combination of local expert decision-makers. Approach (ii) leads to decision-makers specialized in solving families of tasks, which equips the system with the ability to solve meta-learning problems. We show the broad applicability of our approach on a range of problems including classification, regression, density estimation, and reinforcement learning problems, both in the standard machine learning setup and in a meta-learning setting. 
\end{abstract}

\section{Introduction}
\label{sec:intro}
Intelligent agents are often conceptualized as decision-makers that learn probabilistic models of their environment and optimize utilities or disutilities like cost or loss functions \cite{VonNeumann2007}. In the general case we can think of a utility function as a black-box oracle that provides a numerical score that rates any proposed solution to a supervised, unsupervised or reinforcement learning problem. In context of decision-making, na\"{i}vely enumerating all possibilities and searching for an optimal solution is usually prohibitively expensive. Instead, intelligent agents must invest their limited resources in such a way that they achieve an optimal trade-off between expected utility and resource costs in order to enable efficient learning and acting. This trade-off is the central issue in the fields of bounded or computational rationality with repercussions across other disciplines including economics, psychology, neuroscience and artificial intelligence \cite{payne1993adaptive,Simon1955,aldrich1999organizations,manson2006bounded,gigerenzer2009homo,damasio2009neuroscience,Gershman2015,Genewein2015,Hihn2018}. The information-theoretic approach to bounded rationality is a particular instance of bounded rationality where the resource limitations are modeled by information constraints \cite{Edward2014,McKelvey1995,Tishby2011,wolpert2006information,Ortega2011a,Ortega2013,gottwald2019bounded,Schach2018,lindig2019analyzing} closely related to Jaynes' maximum entropy or minimum relative entropy principle \cite{jaynes1996probability}.

At the heart of information-theoretic models of bounded rationality lies the information utility trade-off for lossy compression, abstraction and hierarchy formation \cite{Genewein2015}. The optimal utility information trade-off leads to an optimal arrangement of decision-makers and encourages the emergence of specialized agents which in turn facilitates an optimal division of labor reducing computational effort \cite{Hihn2018,Gottwald2019}.

In the context of machine learning, hierarchical inference can also be regarded as an example of bounded rational decision-making with information constraints, where different models correspond to different experts that are bound by their respective priors over parameters and try to optimize the marginal likelihood given data \cite{Genewein2015}. The priors that are acquired by the different expert models can be regarded as
efficient abstractions that facilitate generalization to novel problems.  Hierarchical inference models have been used successfully, for example, to model
 human behavior when learning with very limited resources from a handful of examples while excelling at adapting quickly \cite{jankowski2011meta,genewein2015structure,braun2010structure,kemp2007learning}.  
We can study sample-efficient adaptation to new problems as an instance of ``learning to learn'' or meta-learning \cite{thrun2012learning,schmidhuber1997shifting,caruana1997multitask} which is an ongoing and active field of research \cite{koch2015siamese,vinyals2016matching,Finn2017model,ravi2017optimization,ortega2019meta,botvinick2019reinforcement,yao2019hierarchically}.
While we can define meta-learning in different ways \cite{lemke2015metalearning,vilalta2002perspective,giraud2008metalearning,brazdil2008metalearning,hutter2019automated}, a common denominator is that  systems capable of meta-learning happen to learn on two levels, each with a different time scale: slow learning across different tasks (the meta-level), and fast learning to adapt to each task individually. Understanding efficient meta-learning still poses a challenging machine learning problem, as applying pre-trained models to new tasks na\"{i}vely usually leads to poor performance, as with each new incoming batch of data the agent has to perform expensive and slow re-learning of its policy. 

\begin{figure}
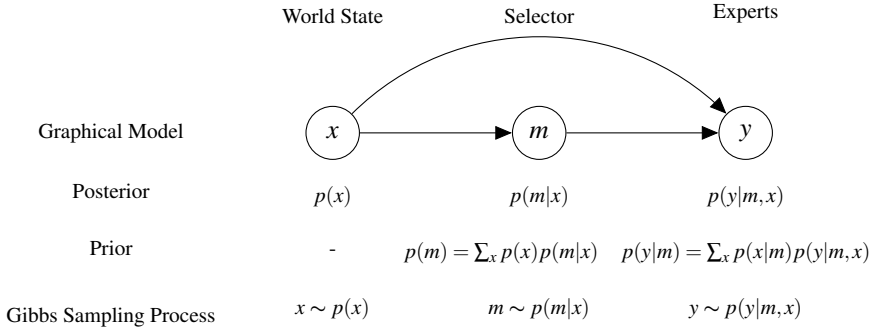

\begin{center}
  \tikz{
 \node[latent] (x) {$x$};
 \node[latent, right=of x, xshift=1cm] (m) {$m$};
 \node[latent, right=of m, xshift=1cm] (y) {$y$};
 
 \node[left=of x, xshift=-0.5cm] (label1) {Graphical Model};
 
 \node[above=of x] (ws) {World State};
 \node[above=of m] (s) {Selector};
 \node[above=of y] (e) {Experts};
 
 \node[below=of x, yshift=0.75cm] (px) {$p(x)$};
 \node[below=of m, yshift=0.75cm] (pm) {$p(m|x)$};
 \node[below=of y, yshift=0.75cm] (py) {$p(y|m,x)$};
 \node[below=of label1, yshift=0.65cm] (label2) {Posterior};
 
 \node[below=of px, yshift=0.65cm] (p0x) {-};
 \node[below=of pm, yshift=0.75cm, xshift=-0.5cm] (p0m) {$p(m) = \sum_x p(x) p(m|x)$};
 \node[below=of py, yshift=0.75cm] (p0y) {$p(y|m) = \sum_x p(x|m)p(y|m,x)$};
 \node[below=of label2, yshift=0.65cm] (label3) {Prior};
 
 \node[below=of px] (gx) {$x \thicksim p(x)$};
 \node[below=of pm] (gm) {$m \thicksim p(m|x)$};
 \node[below=of py] (gy) {$y \thicksim p(y|m,x)$};
 \node[below=of label3, yshift=0.5cm] (label4) {Gibbs Sampling Process};

 \edge {m}{y}; \edge{x}{m};
 \draw [->] (x) to [out=45,in=135] (y);
 }
 \end{center}
 \caption{The hierarchical expert model: after observing a world state $x$, the selector samples an expert $m$ according to a selection policy $p(m|x)$ and then the expert samples an action $y$ from the expert's posterior action policy $p(y|m,x)$. This can be seen as a Gibbs sampling process. Each posterior is the result of a trade-off between maximizing the utility and minimizing the $\DKL$ to the respective prior.}
 \label{fig:infmodel}
\end{figure}

In this study, we aim to harness the efficient adaptation and the ability of meta-learning of hierarchical inference processes  for the learning of decision-making hierarchies formulated in terms of arbitrary utility functions -- see Figure \ref{fig:infmodel} and Table \ref{tab:setup}. The formulation in terms of general utility functions makes this approach applicable to a broad range of machine learning problems that can be formulated as optimization problems, including supervised and unsupervised learning, and reinforcement learning.
To this end, we extend our previous work on specialization in hierarchical decision-making systems introduced in Hihn et al.\, (2019) \cite{hihn2019} to the problem of meta-learning. After introducing information-theoretic constraints for learning and decision-making in Section~\ref{sec:background}, 
we explain our hierarchical online learning approach to classification, regression, unsupervised learning and reinforcement learning in Section~\ref{sec:spec} for the case of \textit{within task} specialization. We extend our mixture-of-experts learning experiments from Hihn et al.\, (2019) \cite{hihn2019} for supervised and reinforcement learning in Sections~\ref{sec:experimentsclass} and \ref{sec:experimentsrl} and devise a novel application to density estimation in Section~\ref{sec:experimentsunsuper}. The extended experiments in the classification and reinforcement learning setting provide new insights into how the number of experts influences the information processing and classification error and how expert policies partition the reinforcement learning problem space amongst themselves.
In Sections~\ref{sec:specmeta} and \ref{sec:experimentsmeta} we extend the approach from state-based specialization to the case of \textit{across task}  specialization. We show that this task specialization gives rise to a novel meta-learning approach where the meta-learner  assigns previously unseen tasks to experts specialized on similar tasks. In order to split the task space and to assign the partitions to experts, we learn to represent tasks through a common latent embedding, which a gating network uses to distribute tasks among the experts--similar to the posterior over models in a hierarchical inference setup. In Section~\ref{sec:disc}, we discuss novel aspects of the current study in the context of previous literature and conclude with a final summary.

\section{Information Constraints in Learning and Decision-making}
\label{sec:background}

\subsection{Decision-making with information constraints}
\label{seq:br}
Given a utility function $\mathbf{U}(x, y)$ that indicates the desirability of each action $y \in \altmathcal{Y}$ taken in each context $x \in \altmathcal{X}$, a fully rational agent picks action $y^*_{x} = \argmax_{y} \mathbf{U}(x, y)$. A bounded rational decision-maker with information constraints
 \cite{Ortega2013} is modeled by an upper bound $B$ on the Kullback-Leibler divergence $\DKL(p(y\vert x)\vert\vert p(y)) = \sum_{y}{p(y\vert x) \log{\frac{p(y\vert x)}{p(y)}}}$ between the agent's prior $p(y)$ and  posterior policy $p(y\vert x)$ to express the limitation on the decision-maker's information processing resources for reducing uncertainty when optimizing $\mathbf{U}(x, y)$. This results in the following optimization problem:
\begin{equation}
\max_{p(y\vert x)} \mathbb{E}_{p(y\vert x)}\left[\mathbf{U}(x, y)\right] - \frac{1}{\beta}\mathbb{E}_{p(x)}\left[\DKL(p(y\vert x)\vert \vert p(y))\right].
\label{eq:br_obj}
\end{equation}
by introducing a Lagrange multiplier  $\beta \in \mathbb{R}^+$ that is determined by $B$. 
For $\beta \rightarrow \infty$ we recover the maximum utility solution and for $\beta \rightarrow 0$ the agent can only act according to the prior. In the case of a known state distribution $p(x)$, the optimal prior is given by the marginal $p(y) = \sum_{x}{p(x) p(y\vert x)}$ and the expected Kullback-Leibler divergence becomes equal to the mutual information $I(X;Y)$.

When aggregating bounded-rational agents into hierarchical decision-making systems that split the action space into soft partitions \cite{Genewein2015}, an expert selection policy $p(m\vert x)$ can be introduced that selects an expert $m$ for a given state $x$ that chooses their action according to a stochastic policy $p(y\vert x,m)$. Similar to Equation~\eqref{eq:br_obj}, such a hierarchical decision-making system with information constraints can be expressed by the following optimization problem:   
\begin{equation}
\label{eq:par_mutual}
\max_{p(y\vert x,m), p(m\vert x)} \E[\mathbf{U}(x,y)] - \frac{1}{\beta_1}I(X;M) - \frac{1}{\beta_2}I(X;Y\vert M),
\end{equation}
where $\beta_1$ is the resource parameter for the expert selection stage and $\beta_2$ for the experts, and $I(\cdot;\cdot|M)$ denotes the conditional mutual information. The optimal solution is a set of coupled equations
\begin{equation}
\label{eq:pareqs}
\begin{cases}
\begin{array}{rcl}
p(m\vert x) &=& \frac{1}{Z(x)}p(m) \exp(\beta_1 \F(x,m)) \\[2pt]
p(y\vert x,m) &=& \frac{1}{Z(x,m)} p(y\vert m) \exp(\beta_2 \mathbf{U}(x,y)) \\[2pt] 
\end{array}
\end{cases}
\end{equation}
with the marginals $p(m) =\sum_{x} p(x) p(m\vert x)$ and $p(y\vert m) =\sum_{x} p(x\vert m)p(y\vert x,m)$, and the free-energy difference 
$\F(x,m) = \E_{p(y|x,m)}[\mathbf{U}] - \frac{1}{\beta}\DKL(p(y|x,m)\vert\vert p(y|m))$. 
If we assume $\mathbf{U}(x,y)$ to represent the log-likelihood and $m$ to represent different statistical models with parameters $y$ to explain observables $x$, then Equation~\eqref{eq:par_mutual} describes the problem of hierarchical inference in terms of an optimization problem to produce the posteriors $p(y|x,m)$ and $p(m|x)$.

\subsection{Learning with Information Constraints}
In learning problems, the true utility $\mathbf{U}(x, y)$ is unknown, instead we are given a function $\mathbf{U}_D(x, y)$ after having observed $n$ data samples $D = \{(x_i,y_i)\}_{i=1}^n$ and we know that $\mathbf{U}_D(x, y)\rightarrow\mathbf{U}(x, y)$ for $n\rightarrow\infty$. Assuming that we know $p(x)$, we are looking for the best strategy $p(y|x)$ that optimizes $\mathbf{U}(x, y)$. If we were to treat this as an inference problem, we could index different candidate solutions $p(y|x,\theta)$ with a parameter $\theta$ and place a prior $\pi(\theta)$ over this parameter. From PAC-Bayes analyses it is well-known that the Gibbs distribution $\pi_D(\theta)\propto \pi(\theta)\exp\left(\gamma \mathbf{U}_D(\theta)\right)$ minimizes the generalization error \cite{mcallester2003pac,mcallester1999pac}
\[
\mathbb{E}_{\pi_D(\theta)}\left[\mathbf{U}(\theta) - \mathbf{U}_D(\theta)\right] \leq \sqrt{\frac{\DKL(\pi_D(\theta)||\pi(\theta)) + \log\frac{n}{\delta} }{2(n-1)} }
\]
where $U_n(\theta) = \frac{1}{n}\sum_{x \in D}\sum_y p(y|x,\theta) \mathbf{U}_D(x, y)$ and $U(\theta) = \sum_{x,y} p(x) p(y|x,\theta) \mathbf{U}(x, y)$. 
In a classification problem, for example, we could choose the utility $\mathbf{U}(x,y)=\mathbb{I}_{y=h(x)}$ where $\mathbb{I}$ is the indicator function and $h(x)$ is the mapping from $x$ to the true labels $y_{true}$.
Just like the prior $\pi(\theta)$ can be seen to regularize the search for the parameter $\theta$, we can therefore regard the $\DKL$ as a principled regularizer in the space of probability distributions over $\theta$. Regularization techniques \cite{kukavcka2017regularization,srivastava2014dropout,ioffe2015batch} are typically introduced to overcome the problem of overfitting (i.e.,~ large generalization error), such that we can regard information constraints in $\theta$-space  as a particular instance of regularization in the space of hypotheses.

Instead of studying information constraints  in hypothesis space governing the update from prior $\pi(\theta)$ to posterior $\pi_D(\theta)$, we could also directly study information constraints in the output space governing the update between the prior and posterior predictive distributions $p(y|x) = \sum_\theta \pi(\theta)p(y|x,\theta)$  and $p_D(y|x) = \sum_\theta \pi_D(\theta)p(y|x,\theta)$.
If the update from $\pi(\theta)$ to $\pi_D(\theta)$ is bound by $\DKL(\pi_D(\theta)||\pi(\theta))\leq C_1$, then the update from $p(y|x)$ to $p_D(y|x)$ will be bound by $\DKL(p_D(y|x)||p(y|x))\leq C_2$. This suggests that one could try to use the $\DKL$ in output space directly as a regularizer. Instead of limiting our search for distributions $p(y|x)$ by imposing a prior in $\theta$-space, we limit our search for $p(y|x)$ directly through a prior $p(y)$ for some suitable $C_2$.

Such output regularization techniques have indeed been proposed in the literature. In an unregularized classifier, for example, the probabilities assigned to incorrect class labels would often be pushed close to zero without $\DKL$ regularization, such that $p_{\theta}(y|x)$ collapses to a delta function over the correct class label, which is a sign of overfitting \cite{szegedy2016rethinking}. To avoid this, it has been suggested \cite{pereyra2017regularizing} to view the probabilities assigned to incorrect labels as knowledge the learner has extracted from the dataset, which can be achieved by encouraging the learner to produce output distributions that balance high entropy and minimal loss:
\begin{equation}
\label{eq:confpen}
\theta^* = \argmin_\theta \sum \altmathcal{L}(x, y)- \frac{1}{\beta}H(p_\theta(y|x)),
\end{equation}
where $x,y$ are training data, $\altmathcal{L}(x, y)$ is a error function and $H(p) = -\sum_{x_\in\altmathcal{X}}p(x)\log p(x)$ is the entropy of $p$. This technique introduced by Pereyra et al.~ (2017) \cite{pereyra2017regularizing} has immediate connections to our approach through the following observation: adding the $\DKL$ between the agent's posterior $p_\theta(y|x)$ and prior $p(y)$ recovers confidence penalty, if the agent's prior policy is uniform.
A similar idea has also been suggested in the context of using label smoothing as an output regularization technique \cite{muller2019does}.

In reinforcement learning encouraging the agent to learn policies with high entropy is a widely applied technique known as maximum entropy reinforcement learning (RL) \cite{haarnoja2017reinforcement}. Maximum entropy RL typically penalizes deviation from a fixed uniformly distributed prior to promote exploration, but in a more general setting we can discourage deviation from an arbitrary prior policy by optimizing for
\begin{equation}
\label{eq:maxentrl}
\max_{p}\mathbb{E}_{p}\left[\sum_{t=0}^\infty\gamma^t \left( r(x_t, a_t) -  \frac{1}{\beta} \log \frac{p(a_t\vert x_t)}{p(a)}\right)\right],
\end{equation}
where $\beta$ trades off between reward and entropy, such that $\beta \rightarrow \infty$ recovers the standard RL value function and $\beta \rightarrow 0$ recovers the value function under a random policy. While the entropy term is often regarded as a simple method to encourage exploration, we could similarly regard it as an information constraint for regularization to prevent overfitting as in the case of supervised learning.

\section{Within Task Specialization in Hierarchical Multi-Agent Policies} 
\label{sec:spec}
\begin{table}[t!]
\centering
\begin{tabular}{*5c}
\toprule
\multicolumn{5}{c}{\normalsize Learning Setup and Variables}  \\
\toprule
{} & Utility & $x$ & $m$ & $y$ \\
\midrule
Supervised Learning & neg. MSE / XEnt & data & expert index & label \\
\midrule
Unsupervised Learning & log-likelihood & data & model index  & parameters \\
\midrule
Reinforcement Learning & reward & state & policy index & action \\
\bottomrule
\end{tabular}
\caption{Our method exhibits flexibility in that we can address diverse problems by defining different utility functions and expert networks. Note that in the Meta-Learning setup $x$ is a Dataset $D$ defining a task and $m$ is an expert over tasks. To assign datasets we compute feature representations $h(D)$, as we illustrate in Figure \ref{fig:features} in the Appendix.}
\label{tab:setup}
\end{table}

In the following we introduce the building blocks of our novel gradient based algorithm to learn the components of a hierarchical multi-agent policy with information constraints. In particular, we leverage the hierarchical model introduced earlier to learn a utility driven partitioning of the state and action spaces. We will demonstrate experimentally how limiting the amount of information each agent can process leads to specialization. First we will show how to transform this principle into a general on-line learning algorithm and afterwards we will derive applications to supervised, unsupervised, and reinforcement learning. In Table \ref{tab:setup} we summarize the different setups in supervised, unsupervised and reinforcement learning with according utility functions.

The model consists of two stages: an expert selection stage followed by an action selection stage. The first stage learns a soft partitioning of the state space and assigns each partition optimally to the experts according to a parametrized stochastic policy $p_\theta(m\vert x)$ with parameters $\theta$ such that under an information-theoretic constraint we can maximize the free energy $\F(x,m)$. We start by rewriting Equation~\eqref{eq:par_mutual} as:

\begin{equation}
\label{eq:overall_objective}
\max_{p_\vartheta(y\vert x,m), p_\theta(m\vert x)} \sum_{x,m,y}p(x)  p_\theta(m\vert x) p_\vartheta(y\vert x,m)J(x,m,y)
\end{equation}
where we define the objective $J(x,m,y)$ as
\begin{equation}
J(x,m,y) = \mathbf{U}(x,y) - \frac{1}{\beta_1}\log \frac{p_\theta(m\vert x)}{p(m)} - \frac{1}{\beta_2}\log \frac{p_{\vartheta}(y\vert x,m)}{p(y \vert m)},
\end{equation}
and $\theta, \vartheta$ are the parameters of the selection policy and the expert policies. Note that each expert policy has a distinct set of parameters $\vartheta = \{\vartheta_m\}_m$, but we drop the $m$ index for readability. 

As outlined in Section \ref{seq:br}, the optimal prior to find an  optimal utility information trade-off is the marginal of the posterior policy given by $p(y) = \sum_{x} p(x) p(y\vert x)$. It would be prohibitive to compute the prior in each step, as it would require marginalizing over large action and state spaces. Instead, we approximate $p(m)$ and $p(y\vert m)$ by exponential running mean averages of the posterior policies with momentum terms $\lambda_1$ and $\lambda_2$:
\begin{eqnarray}
p_{t+1}(y\vert m) &=& \lambda_1p_t(y\vert m) + (1 - \lambda_1)p_\vartheta(y\vert x,m) \\
p_{t+1}(m) &=& \lambda_2p_t(m) + (1 - \lambda_2)p_{\theta}(m\vert x).
\end{eqnarray}

\subsection{Specialization in Supervised Learning}
\label{sec:secsuper}
We set the negative loss as the utility $\U(y, \hat{y}) = -\altmathcal{L}(y, \hat{y})$, where $\hat{y}$ represents the expert's response (predicted class label, regressed value) and $y$ is the true label. In our implementation we use the cross-entropy loss $\altmathcal{L}(y, \hat{y}) = \sum_i y_i \log \frac{1}{\hat{y}_i} = -\sum_i y_i \log \hat{y}_i$ as a  performance measure for classification tasks, and for regression the mean squared error $\altmathcal{L}(y, \hat{y}) = \sum_i \vert\vert \hat{y}_i-y_i\vert\vert^2_2 $ between the prediction $\hat{y}$ and the ground truth values $y$. The selection policy thus optimizes
\begin{equation}
\label{eq:supervisedselectorobj}
\max_\theta \E_{p_\theta(m\vert x)}\left[\hat{f}(m,x) - \frac{1}{\beta_1}\log\frac{p_\theta(m\vert x)}{p(m)}\right],
\end{equation}
where $\hat{f}(m,x) \coloneqq \mathbb{E}_{p_\vartheta(\hat{y}\vert m,x)}\big[-\altmathcal{L}(\hat{y},y) - \frac{1}{\beta_2}\log\frac{p_\vartheta(\hat{y}\vert x,m)}{p(\hat{y}\vert m)}\big]$ is the free energy of expert $m$. Note that this introduces a double expectation, which we can estimate by Monte Carlo sampling. The experts thus simply optimize their free energy objective defined by
\begin{equation}
\max_\vartheta \E_{p_\vartheta(\hat{y}\vert m,x)}\left[-\altmathcal{L}(y, \hat{y}) - \frac{1}{\beta_2}\log\frac{p_\vartheta(\hat{y}\vert x,m)}{p(\hat{y}\vert m)}\right].
\end{equation} 

\subsection{Specialization in Unsupervised Learning}
\label{sec:secunsuper}
Unsupervised learning \cite{barlow1989unsupervised} deals with learning from data in the face of missing labels, such as clustering \cite{xu2008clustering} and density estimation \cite{silverman2018density} algorithms. The density estimation method we propose is similar to RBF Networks \cite{schwenker2001three} or Gaussian Mixture Models \cite{biernacki2000assessing}. In the following we will show how our method can handle unsupervised learning, where we interpret each expert as a Normal-Wishart distribution. We model this by learning a distribution $p(\bm{\mu},\bm{\Lambda} \vert \bm{\omega}, \lambda, \bm{W}, \nu)$ over means $\bm{\mu}$ and covariance matrices $\bm{\Lambda}$ as Normal-Wishart distributions:
\begin{equation}
p(\bm{\mu},\bm{\Lambda} \vert \bm{\omega}, \lambda, \bm{W}, \nu) = \altmathcal{N} \left(\bm{\mu} \vert \bm{\omega}, (\lambda \bm{\Lambda})^{-1} \right) \altmathcal{W}(\bm{\Lambda} \vert \bm{W}, \nu),
\end{equation}
where $\altmathcal{W}$ is Wishart a distribution, $\altmathcal{N}$ a Normal distribution, $\bm{\omega} \in \mathbb{R}^D$ is the mean of the normal distribution, $\bm{W} \in \mathbb{R}^{D\times D}$ is the scale matrix, $\nu > D - 1$ is the degree of freedom, $\lambda > 0$ is a scaling factor, and $D$ denotes the dimensionality of the data. Sampling is straightforward: we first sample $\bm{\Lambda}$ from a Wishart distribution with parameters $\mathbf{W}$ and $\nu$. Next we sample $\bm{\mu}$ from a multivariate normal distribution with mean $\bm{\omega}$ and variance $(\lambda \bm{\Lambda })^{-1}$. We assume the data $x$ follows a normal distribution $x \thicksim \altmathcal{N}(\bm{\mu}, (\lambda \bm{\Lambda})^{-1})$. The parameters $\nu$, $\lambda$ are hyper-parameters we set beforehand, such that we are interested in finding the parameters $\bm{\mu}^*$  and $\bm{W}^*$ maximizing the likelihood of the data:
\begin{equation}
\omega^*, \bm{W}^* = \argmax_{\bm{W}, \omega} \altmathcal{N}(x|\mu, (\lambda\bm{\Lambda})^{-1})p(\mu,\bm{\Lambda}|\omega,\bm{W},\lambda,\nu)
\end{equation}
Thus, in this setting the expert's task is to find parameters $\omega^*$ and $\bm{W}^*$ in order to select a tuple $(\bm{\mu}, \bm{\Lambda})$ that models the likelihood of the data well. The objective of the selector is to assign data to the experts that not only have a set of parameters that yield high likelihood on the assigned data, but also have low statistical complexity as measured by the $\DKL$ between the expert's posterior and prior distributions. We can now define the free energy difference for each expert as
\begin{equation}
\hat{f}(x,m) = \mathbb{E}_{p_\vartheta(\bm{\mu},\bm{\Lambda}\vert m)}\left[\ell(x\vert \bm{\mu}, (\lambda \bm{\Lambda})^{-1}) - \frac{1}{\beta_2}\DKL(p(\bm{\mu},\bm{\Lambda})\vert \vert p_0(\bm{\omega}_0,\bm{\Lambda}_0)\right],
\end{equation}
where $p(\bm{\mu},\bm{\Lambda})$ is the expert's posterior Normal-Wishart distribution over the parameters $\mu$ and $\lambda$ and $p_0(\bm{\omega}_0,\bm{\Lambda}_0)$ is the expert's prior, $p$ and $p_0$ are the experts posterior and prior distribution and  $\ell(x\vert \bm{\mu}, (\lambda \bm{\Lambda})^{-1})$ is the Gaussian log likelihood
\begin{equation}
\ell(x\vert \bm{\mu}, (\lambda \bm{\Lambda})^{-1}) =  -\frac{1}{2}\log(\vert (\lambda \bm{\Lambda})^{-1} \vert) + (x - \bm{\mu})^T(\lambda \bm{\Lambda})(x - \bm{\mu}) + D \log(2\pi) 
\end{equation}
of a data point $x$ given the distribution parameters $\bm{\mu}, (\lambda \bm{\Lambda})^{-1}$.
This serves as the basis for the selector's task of assigning data to the expert with maximum free energy by optimizing
\begin{equation}
\max_\theta \E_{p_\theta(m\vert x)}\left[\hat{f}(x,m) - \frac{1}{\beta_1} \log\frac{p_{\theta}(m\vert x)}{p(m)}\right].
\end{equation}
We can compute the $\DKL$ between two Normal-Wishart distributions $p$ and $q$ as
\begin{equation}
\begin{split}
\DKL \left[ p(\bm{\mu}, \bm{\Lambda}) \Vert q(\bm{\mu}, \bm{\Lambda}) \right] = \frac{\lambda_{q}}{2} \left( \bm{\mu}_{q} - \bm{\mu}_{p} \right)^{\top} \nu_{p} \mathbf{W}_{p} \left( \bm{\mu}_{q} - \bm{\mu}_{p} \right) - \\ \frac{\nu_q} 2 \log \vert \bm{W}_q^{-1} \bm{W}_p\vert   + \frac{\nu_p}{2}(\text{tr}(\bm{W}_q^{-1} \bm{W}_p) - D) + C,
\end{split}
\end{equation}
where $C$ is a term that does not depend on the parameters we optimize, so we can omit it, as we are only interested in relative changes in the $\DKL$ caused by changes to $\bm{W}$ and $\bm{\omega}$ (see Appendix \ref{app:wishartdkl} for details on the derivation).

\subsection{Specialization in RL Agents}
\label{sec:specrl}

In reinforcement learning we model sequential decision problems by defining a Markov Decision Process (MDP) as a tuple $(\altmathcal{S}, \altmathcal{A}, P, r)$, where $\altmathcal{S}$ is the set of states, $\altmathcal{A}$ the set of actions, $P: \altmathcal{S} \times \altmathcal{A} \times \altmathcal{S} \rightarrow [0,1]$ is the transition probability, and $r: \altmathcal{S} \times \altmathcal{A} \rightarrow \mathbb{R}$ is a reward function. The aim is to find the parameter $\theta^* = \argmax_{\theta} J(p_\theta)$ of a policy $p_\theta$ that maximizes the expected discounted reward $J(p_\theta)=  \mathbb{E}_{p_\theta}\left[\sum_{t=0}^T \gamma^t r(x_t, a_t)\right]$. In case of an infinite horizon, we have $T\rightarrow\infty$.
We define $r(\tau) = \sum_{t=0}^T \gamma^t r(x_t, a_t)$ as the cumulative reward of trajectory $\tau = \{(x_t, a_t)\}_{i=0}^T$, where we generate the trajectory according to the policy $p(a_t|x_t)$ and the environmental dynamics $P(x_{t+1}\vert x_t, a_t)$.
Reinforcement learning \cite{Sutton2018} models assume that an agent interacts with an environment over a number of discrete time steps $t$. At each time step $t$, the agent finds itself in a state $x_t$ and selects an action $a_t$ according to the policy $p(a_t\vert x_t)$. In return, the environment transitions to the next state $x_{t+1}$ and generates a scalar reward $r_t$. 
Here, we consider policy gradient methods \cite{Sutton2000} which are a popular choice to tackle continuous reinforcement learning problems. The main idea is to directly manipulate the parameters $\theta$ of the policy in order to maximize the objective $J(p_\theta)$ by taking steps in the direction of the gradient $\nabla_\theta J(p_\theta)$. 

In the following we will derive our algorithm for specialization in hierarchical reinforcement learning agents. Note that in the reinforcement learning setup the reward function $r(x,a)$ defines the utility $\mathbf{U}(x,a)$. In maximum entropy RL (see e.g.,\, Haarnoja et al.\, (2017) \cite{haarnoja2017reinforcement}) the regularization penalizes deviation from a fixed uniformly distributed prior, but in a more general setting we can discourage deviation from an arbitrary prior policy by optimizing for: 
\begin{equation}
\label{eq:maxentrl}
\max_{p}\mathbb{E}_{p}\left[\sum_{t=0}^T\gamma^t \left( r(x_t, a_t) -  \frac{1}{\beta} \log \frac{p(a_t\vert x_t)}{p(a)}\right)\right],
\end{equation}
where $\beta$ trades off between reward and entropy, such that $\beta \rightarrow \infty$ recovers the standard RL value function and $\beta \rightarrow 0$ recovers the value function under a random policy. 

To optimize the objective \eqref{eq:maxentrl} we define two separate kinds of value function, $V_\phi$ for the selector and one value function $V_\varphi$ for each expert. Thus, each expert is an actor-critic with separate actor and critic networks. Similarly, the selector has an actor-critic architecture, where the actor network selects experts and the critic learns to predict the expected free energy of the experts depending on a state variable. The selector's policy is represented by $p_\theta$, while each expert's policy is represented by a distribution $p_\vartheta$.

\subsubsection{Value Functions}
In standard reinforcement learning the discounted reward is defined as
\begin{equation}
R_t = \sum_{l=0}^T\gamma^lr(x_{t+l}, a_{t+l}),
\end{equation}
which is usually learned through a parameterized value function $V_\psi$ by regressing
\begin{equation}
\psi^* = \argmin_\psi \frac{1}{\vert \altmathcal{D}\vert T}\sum_{\tau\in\altmathcal{D}}\sum_{t=0}^T (V_{\psi}(x_t) - R_t)^2 .
\end{equation}
Here $\psi$ are some arbitrary parameters of the value representation, $V_\psi(x_t)$ is the predicted value estimate for state $x_t$, and $\altmathcal{D}$ is a set of trajectories $\tau$ up to horizon $T$ collected by roll-outs of the policies. 

Similar to the standard discounted reward $R_t$, we can now define a discounted free energy $F_t$ as
\begin{equation}
F_t = \sum_{l=0}^T\gamma^l f(x_{t+l}, m_{t+l}, a_{t+l}),
\end{equation}
where $f(x,m,a) = r(x, a) - \frac{1}{\beta_2} \log\frac{p_{\vartheta}(a\vert x,m)}{p(a\vert m)}$. Accordingly, we can learn a value function $V_\varphi$ for each expert by parameterizing the value function with a neural network and performing regression on $F_T$.
Similarly, we can define a discounted free energy $\bar{F}_t$ for the selector
\begin{equation}
\bar{F}_t = \sum_{l=0}^T\gamma^l \bar{f}(x_{t+l}, m_{t+l}),
\end{equation}
with $\bar{f}(x,m) = \mathbb{E}_{p_\vartheta(a\vert x,m)}[r(x,a) - \frac{1}{\beta_2} \log \frac{p(a\vert x, m)}{p(a\vert m)}]$ that is learned through the selector's value function $V_\phi$ by regressing $\bar{F}_t$.

\subsubsection{Policy Learning} 
\label{sec:actsel}
In standard reinforcement learning a common technique to update a parametric policy representation $p_\omega(a|x)$ with parameters $\omega$ is to use policy gradients that optimize the  cumulative reward
\begin{equation}
J(\omega) = \mathbb{E}\left[ p_\omega(a|x) V_\psi(x)\right]
\end{equation}
expected under the critic's prediction $V_\psi(x)$, by following the gradient
\begin{equation}
\nabla_\omega J(\omega) = \mathbb{E}\left[\nabla_\omega \log p_\omega(a|x) V_\psi(x) \right] .
\end{equation}
This policy gradient formulation \cite{Sutton2000} is prone to producing high variance gradients. A common technique to reduce the variance is to formulate the updates using the advantage function instead of the reward \cite{arulkumaran2017deep}. The advantage function $A(a_t,s_t)$ is a measure of how well a certain action $a$ performs in a state $x$ compared to the average performance in that state, i.e.,~$A(a,x) = Q(x,a) - V_\psi(x)$. Here, $V(x)$ is the value function and is a measure of how well the agent performs in state $x$, and $Q(x,a)$ is an estimate of the cumulative reward achieved in state $x$ when the agent executes action $a$. Thus, the advantage is an estimate of how advantageous it is to pick $a$ in state $x$ in relation to a baseline performance $V_\psi(x)$. Instead of learning the value and the Q function, we can define the advantage function solely based on the critic's estimate $V_\psi(x)$ in the following way
\begin{equation}
A(x_t,a_t) = \underbrace{r(x_t, a_t) + \gamma V_{\psi}(x_{t+1})}_{\approx Q(x_t, a_t)} - V_{\psi}(x_t),
\end{equation}
giving the following gradient estimates for the policy parameters
\begin{equation}
\omega\leftarrow \omega- \frac{\alpha}{\vert \altmathcal{D}\vert }\sum_{\tau \in \altmathcal{D}}\sum_{t=0}^T\nabla_\omega \log p_\omega(a_t\vert x_t)A(x_t,a_t),
\end{equation}
where $\alpha$ is a learning rate and $\altmathcal{D}$ is a set of trajectories $\tau$ produced by the policies.

Similar to the standard policy update based on the advantage function,
the expert selection stage can be formulated by optimizing the expected advantage $\mathbb{E}_p(a|x,m) \left[A_m(x,a)\right]$ for expert $m$ with
\[
A_m(x_t,a_t) = f(x_t,m,a_t) + \gamma V_\varphi(x_{t+1})-V_\varphi(x_t).
\]

Accordingly, we can define an expected advantage function $\mathbb{E}_p(m|x) \left[\bar{A}(x,m)\right]$ for the selector with
\[
\bar{A}(x,m) = \mathbb{E}_{p_\vartheta(a|x,m)} \left[ A_m(x,a) \right].
\]

We estimate the double expectation by Monte Carlo sampling, where in practice we use a single $(x, m, a)$ tuple for $\hat{f}(x,m)$, which enables us to employ our algorithm in an on-line optimization fashion.

\begin{figure}[t!]
\centering
\includegraphics[width=0.95\textwidth, trim={3.15cm 0cm 3.5cm 0cm}, clip]{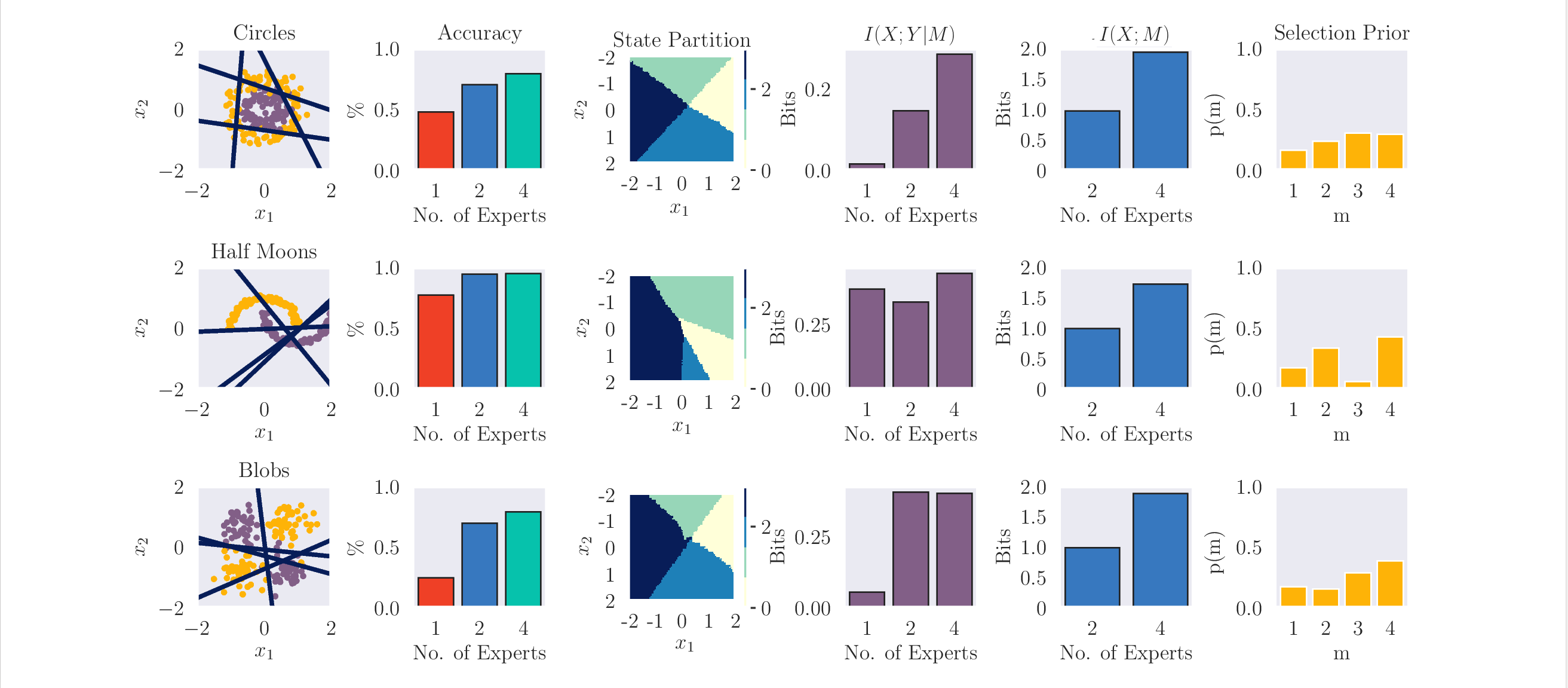}
\caption{Results for three synthetic classification tasks. Our system successfully enables the linear experts to classify their assigned samples correctly by learning a soft partition of the sample space. As expected the accuracy improves and the information processed by each expert increases, as it specializes on the assigned region. We report all quantities presented in this plot on a 20\% hold-out set and averaged over a 10-fold cross-validation scheme.}
\label{fig:classification}
\end{figure}
\begin{figure}[t!]
\centering
\includegraphics[width=0.85\textwidth, trim={1.5cm 0.5cm 1.8cm 1.25cm}, clip]{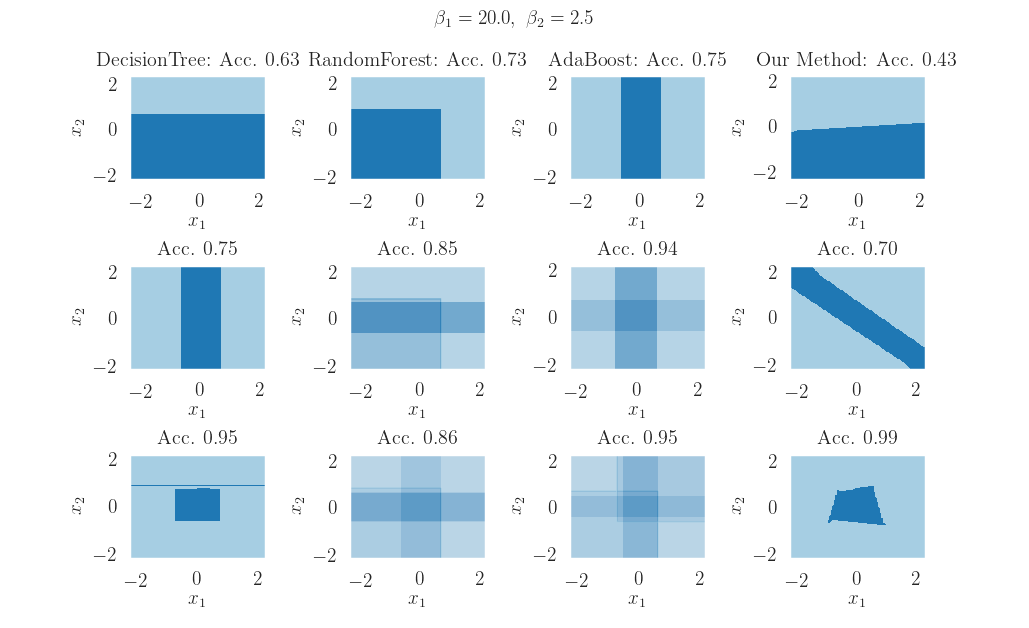}
\caption{Baseline comparison for the "Circles" task. Partitioning of the $x_1 ,x_2$-space is shown for a single decision tree (left column) with depths $1,2,4$, a random forest (middle left column) with $1,2,4$ decision trees of depth $2$, and an Adaboost ensemble (middle right column) of $1,2,4$ decision trees of depth $2$. Our method with $1,2,4$ linear  experts is shown in the rightmost column.}
\label{fig:classificationBaseline}
\end{figure}

\section{Experiments and Results: Within Task Specialization}
\label{sec:experiments}
\begin{figure}
\centering
\includegraphics[width=.45\textwidth, trim={0cm 0cm 1.5cm 0.5cm}, clip]{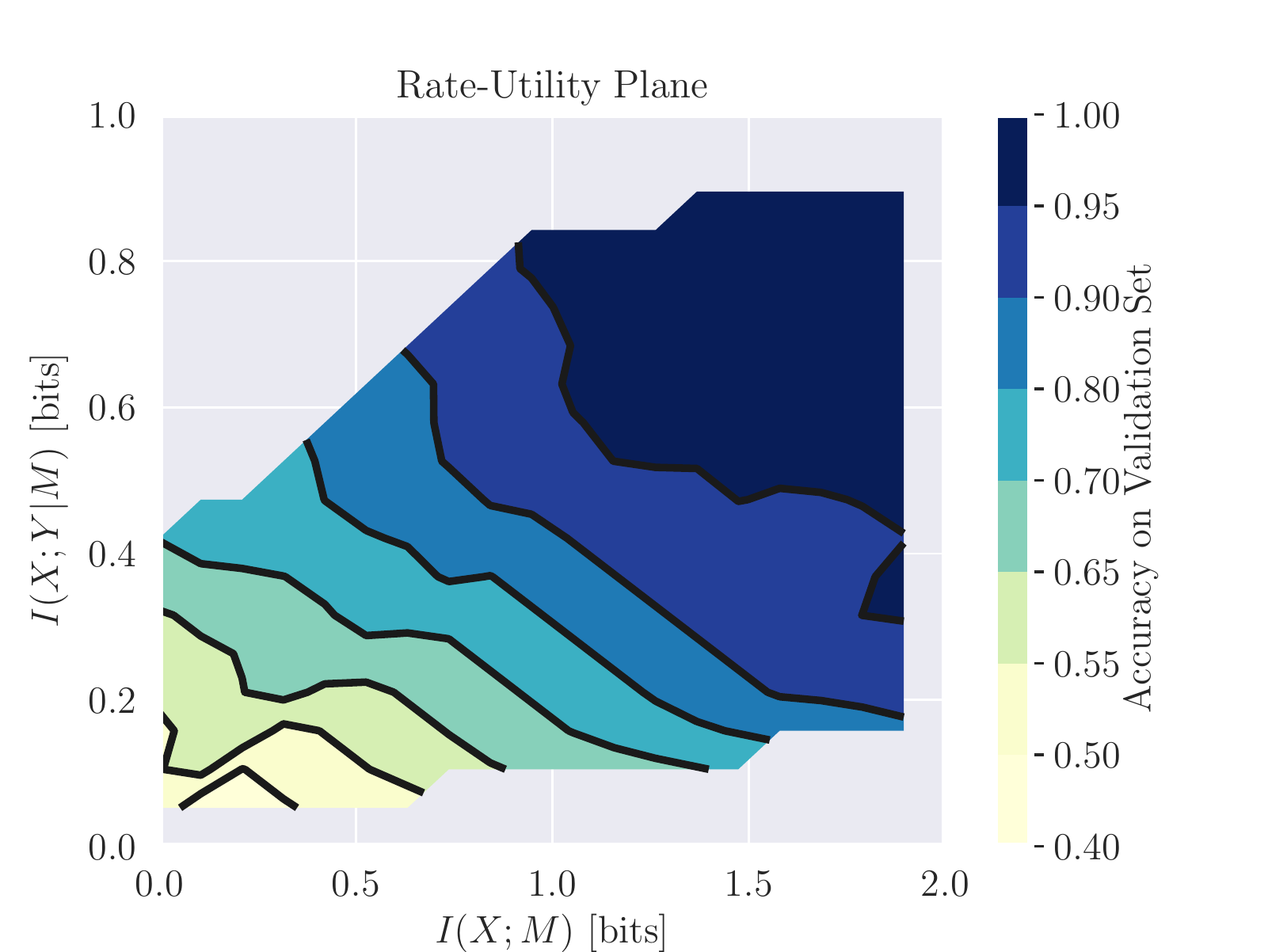}
\includegraphics[width=.45\textwidth, trim={0cm 0cm 1.5cm 0.5cm}, clip]{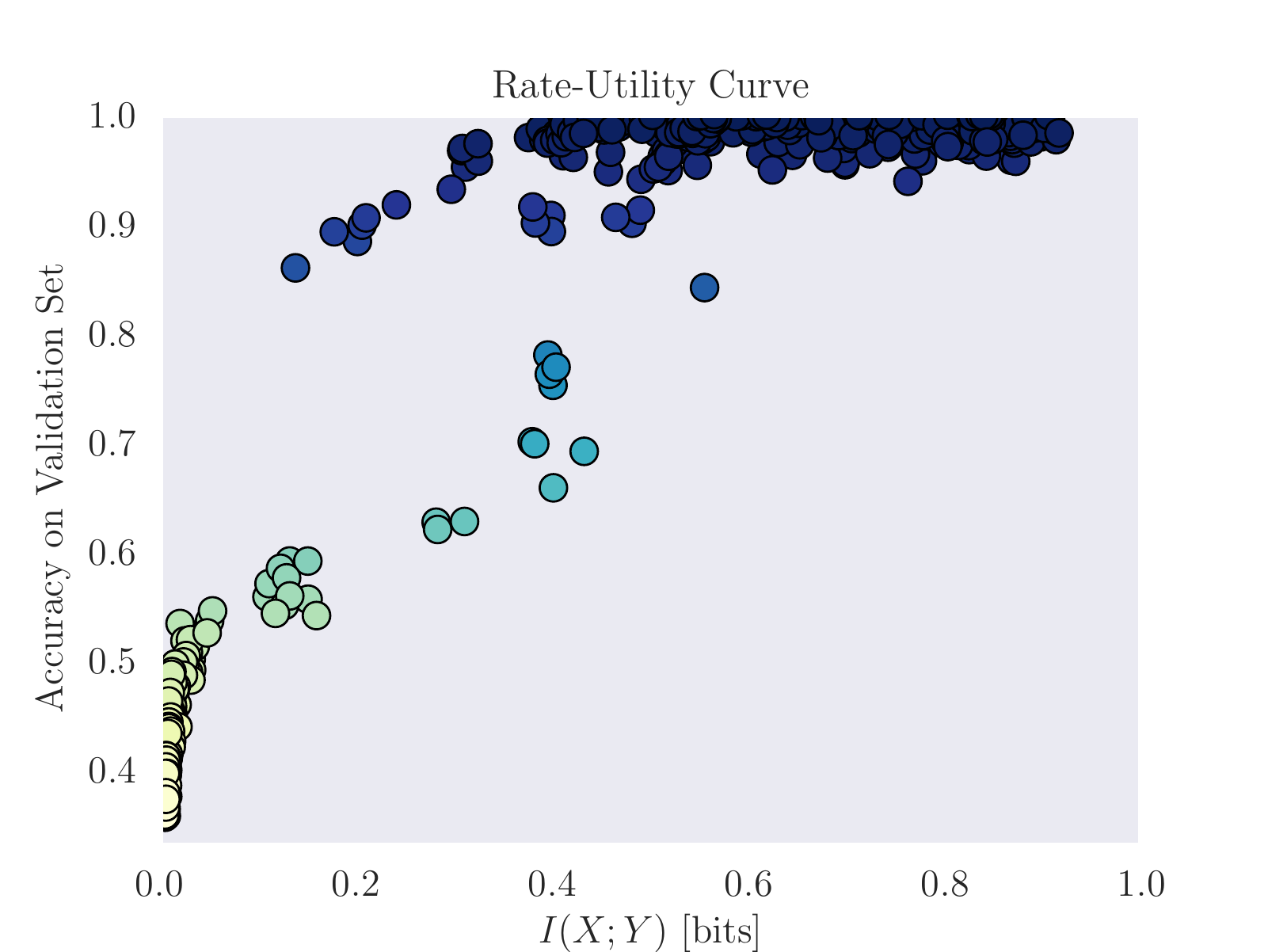}
\caption{Here we show the influence of the information processing in the expert selection stage and in the expert stage (measured by $I(X;M)$ and $I(X;Y\vert M)$) on the classification accuracy synthetic dataset ``circles'' in Figure~\ref{fig:classification}. The grey area is beyond the efficiency frontier of the system. We created this surface by training the system with varying numbers of experts and different settings for $\beta_1$ and $\beta_2$. We approximate intermediate values by a linear interpolation. The right figure shows the information processing of the whole system given by $I(X;Y)$, which we get by marginalizing over the experts. To obtain a particular information processing rate one must choose both parameters correctly, which can be a bit difficult, as we sweep through these parameters in discrete steps. This causes the bumps and irregularly spaced points on the Rate-Utility grid. Some points are almost impossible to reach even though they are theoretically possible, as the convergence of the learning algorithm may suffer under certain $\beta_1, \beta_2$ configurations.}
\label{fig:rateutilityplane}
\end{figure}
In the following we will show applications to different learning tasks where the overall complexity of the problem exceeds the processing power of the individual (linear) experts. In particular, we will look at classification, density estimation, and reinforcement learning. See Appendix \ref{app:experiments} for implementation details. In Section \ref{sec:critical} we discuss how to choose proper values for $\beta_1$ and $\beta_2$ and their effects on learning.

\subsection{Classification with Linear Decision-Makers}
\label{sec:experimentsclass}
When dealing with complex data for classification (or regression) it is often beneficial to combine multiple classifiers (or regressors). In the framework of ensemble learning, for example, multiple classifiers join together to stabilize learning and improve the results \cite{Kuncheva2004}, such as Mixture-of-Experts \cite{Yuksel2012} and Multiple Classifier systems \cite{Bellmann2018}. The method we propose when applied to classification problems falls within this scope of algorithms. The application to regression is an example of local linear regression \cite{atkeson1997locally}.

We evaluate our method on three synthetic datasets for classification--see Figure~\ref{fig:classification}. Our method is able to partition the problem set into subspaces (third column from the left) and fit a linear decision-maker on each subset, achieving acceptable classification accuracy on synthetic non-linear datasets. 
The results on the ''Half Moons'' dataset show an example where the quality of the classification does not improve with the number of experts, essentially because a satisfactory performance can already be achieved with a single expert. We can see that a single expert is classifying the data reasonably well and adding more experts improves accuracy only marginally, whereas in the remaining two datasets the accuracy increases with the number of experts. Regarding the information processed by each expert $I(X;Yvert M)$, a single expert on the ''Half Moons'' achieves a competitive score compared to the system with two and four experts, which in turn results in high accuracy. This also manifests in the selection prior $p(m)$ which shows for this dataset a non-uniform division of labor between the experts. In contrast to this, the results on ''Circles'' and ''Blobs'' show how adding more experts is beneficial if the underlying dataset has a highly non-linear structure. In both settings the information processed by a single expert is close to zero bits and classification accuracy is at chance level. Adding more experts allows for specialization and thus increased processing power $I(X;Y\vert M)$ which in turn achieves higher classification accuracy.

\begin{figure}
\centering
\includegraphics[width=0.33\textwidth, trim={1.5cm 0.1cm 2.8cm 0.8cm}, clip]{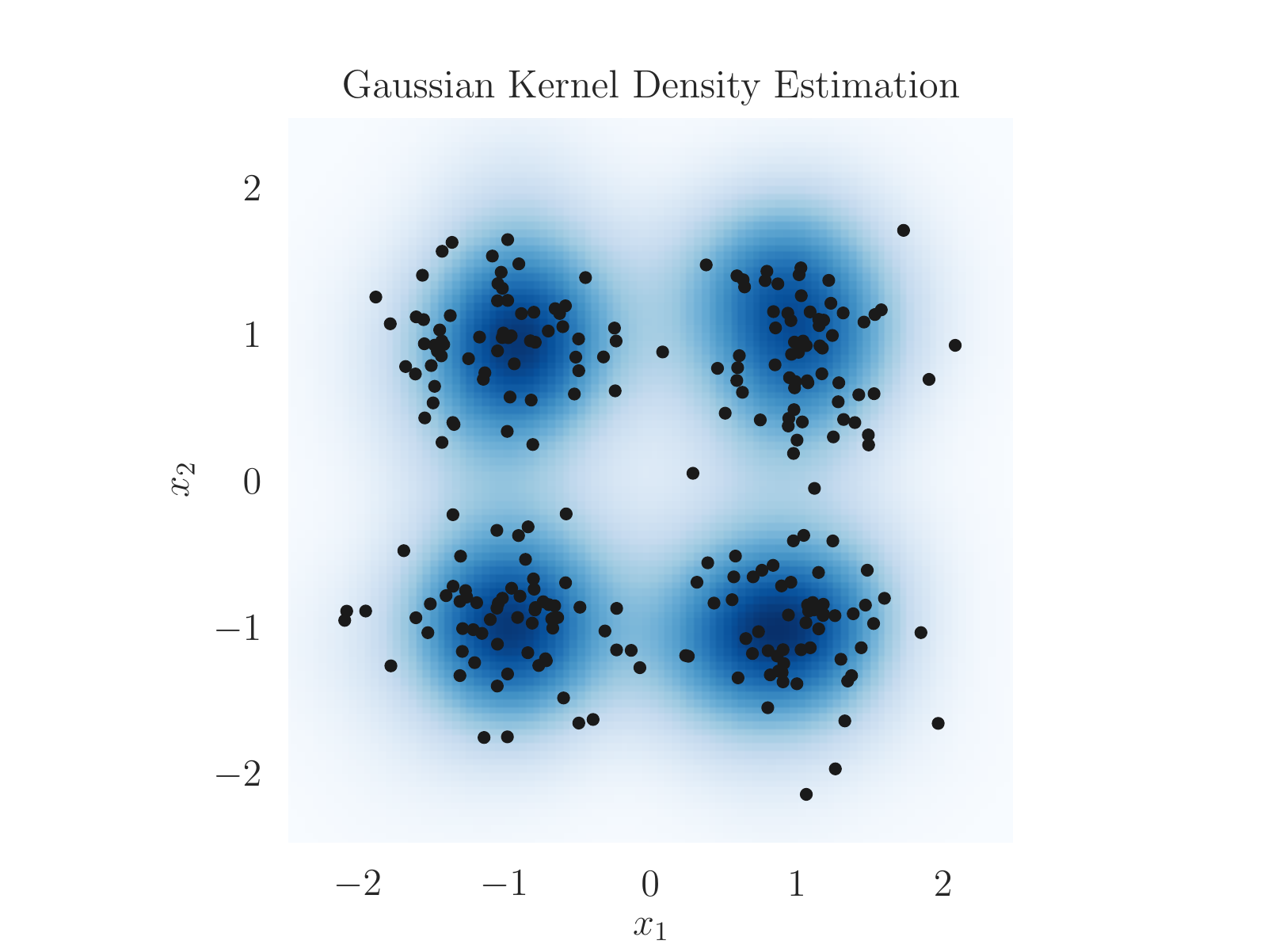}
\includegraphics[width=0.33\textwidth, trim={1.5cm 0.1cm 2.8cm 0.8cm}, clip]{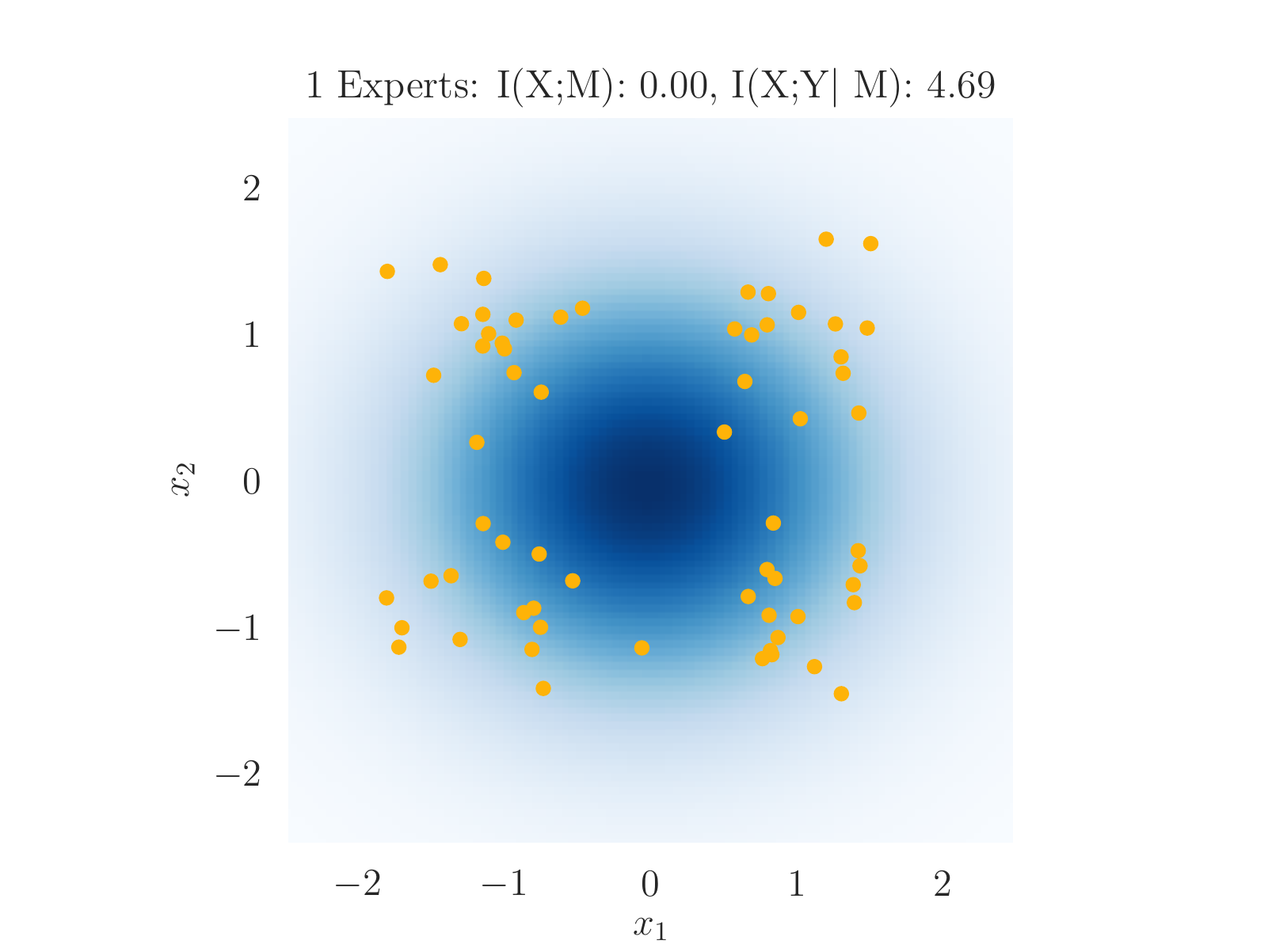}
\includegraphics[width=0.33\textwidth, trim={1.5cm 0.1cm 2.8cm 0.8cm}, clip]{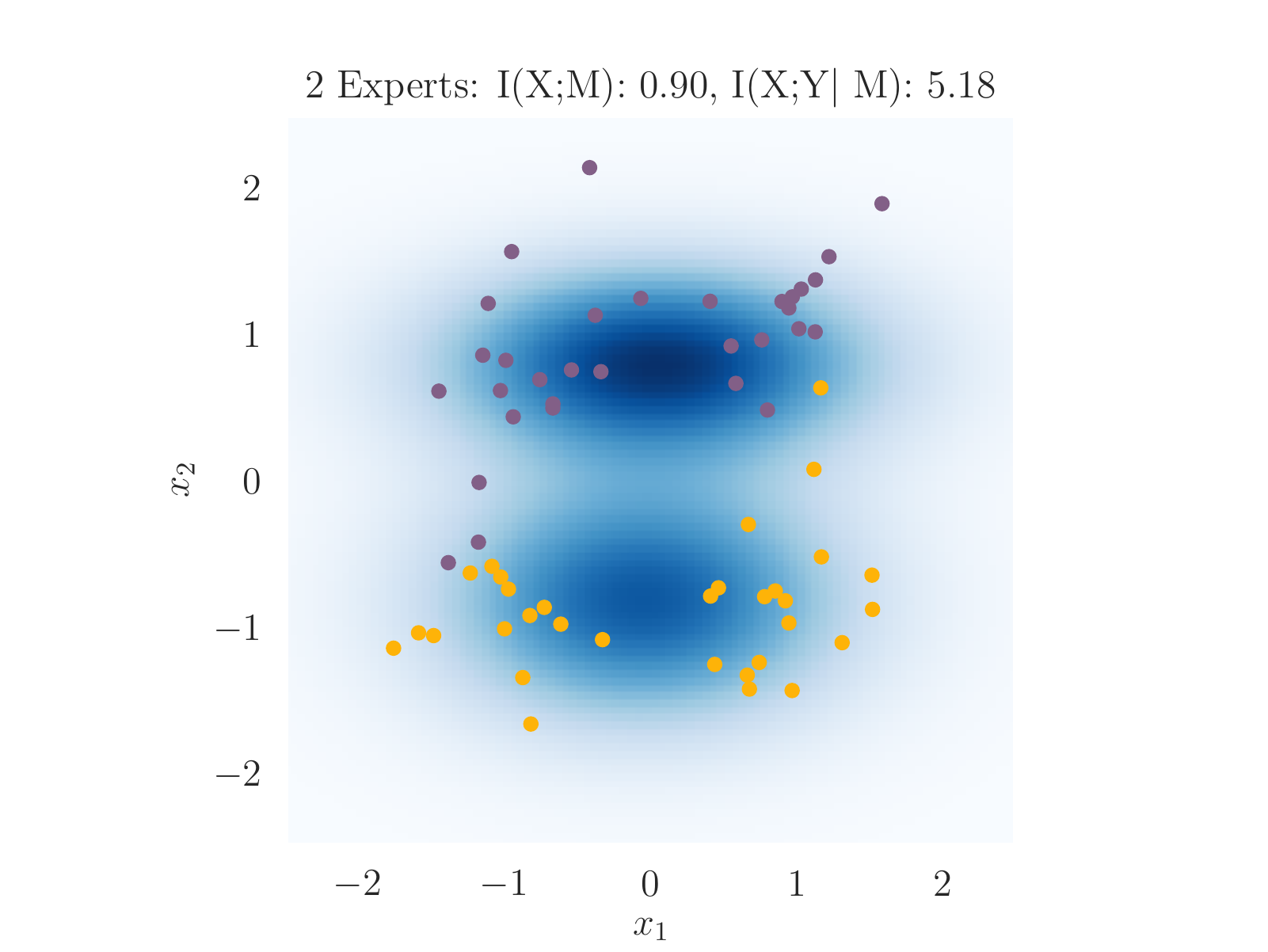}
\includegraphics[width=0.33\textwidth, trim={1.5cm 0.1cm 2.8cm 0.8cm}, clip]{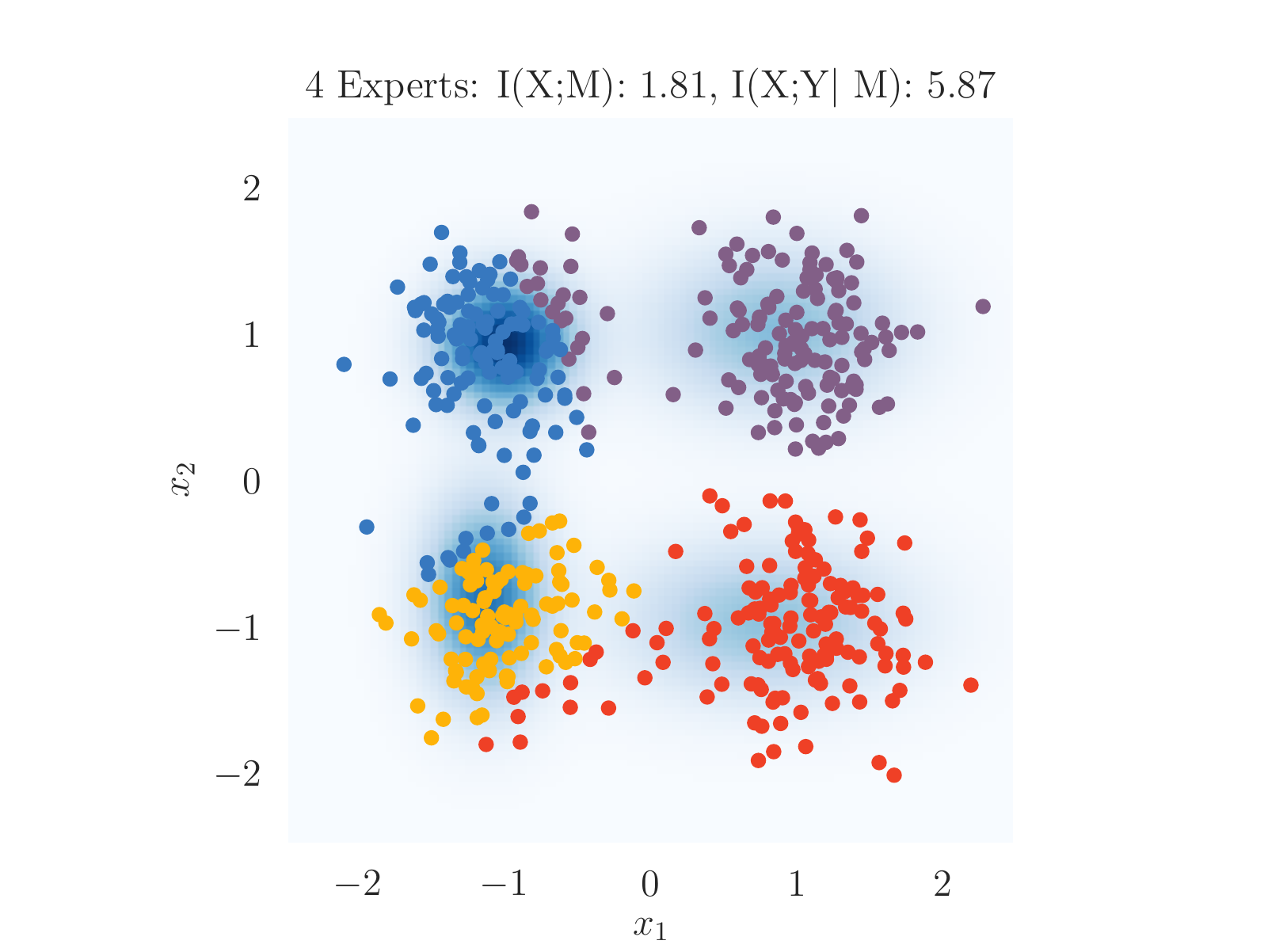}
\caption{Here we show density estimation results on an artificial dataset. We sampled the data from four bivariate Gaussian distributions ($\mu = \{[-1, -1], [-1, 1], [1, 1], [1, 1]\}$ and $\Sigma = 0.15I$) and fit experts on the data. As reference we show the density recovered by gaussian kernel density estimation. Blue indicates high data likelihood. We are able to recover this solution with four experts, but our method allows for additional flexibility by setting the number of experts. As we will show in Section \ref{sec:specmeta}, this abstraction gives rise to meta-learning. We set $\beta_1 = 20$ and $\beta_2 = 1.0$.}
\label{fig:clustering}
\end{figure}

In Figure~\ref{fig:classificationBaseline} we compare the performance of the "Circles" dataset to different baselines constructed from decision tree learning: a single decision tree with varying depth, multiple decision trees as part of a random forest, and multiple decision trees within Adaboost.
In Figure~\ref{fig:rateutilityplane} we report how information processing in the selection and the action stage influences classification accuracy. We can see that under a certain threshold the accuracy is random at best and increases with processing power, which is not surprising. Additionally, we can see that a high selection processing power compensates low processing power in the decision-makers up to a certain degree. If the processing power of the experts is too low, adding more experts does not increase the system's performance indefinitely, as it becomes harder for the selection stage to  pick a certain expert. This happens because  the experts can only specialize on a small subspace of the task due to their low information-processing capabilities.

\subsection{Unsupervised Learning}
\label{sec:experimentsunsuper}
We report unsupervised learning results in Figure~\ref{fig:clustering}, where we show how the system deals with unlabeled data. In this experiment the synthetic dataset contains four clusters and our algorithm is able to perform the clustering as we add more and more experts to partition the state space further. If we provide more experts (in this case 8) than there are clusters (in this case 4), the selector neglects the additional experts and only assigns data to four of them. This indicates that the optimization process aims to find the optimal number of experts in this case.

\subsection{Reinforcement Learning with Linear Decision-Makers}
\label{sec:experimentsrl}
\begin{figure*}[t!]
\centering
\includegraphics[width=0.85\textwidth]{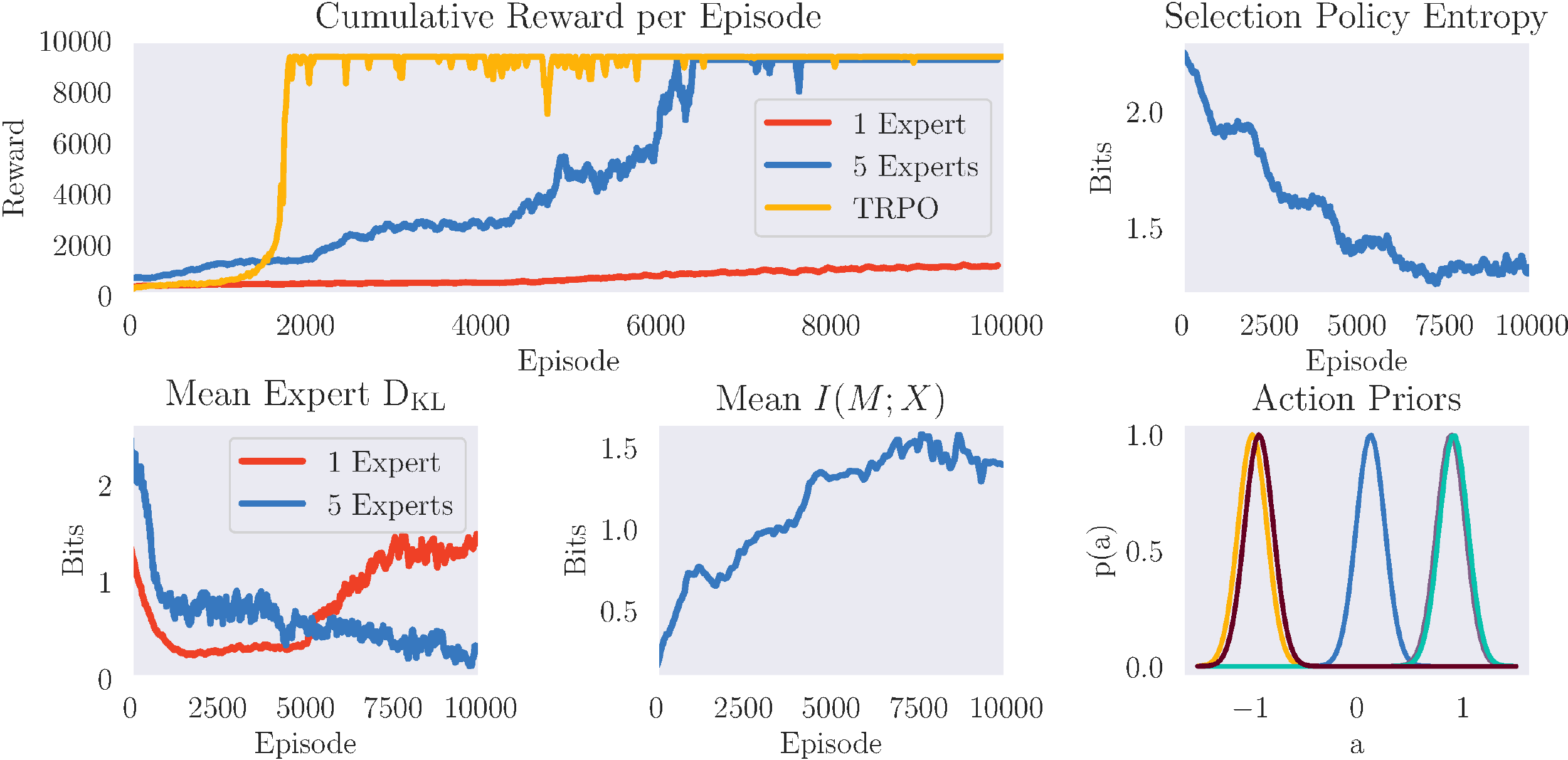}
\caption{Results for the inverted double pendulum problem. The upper row gives the  episodic reward for different system configurations. We show the performance of a system with one linear expert, five linear experts, and compare it to Trust Region Policy Optimization (TRPO) \cite{Schulman2015} (discussed further in Section \ref{sec:relatedwork}). We set $\beta_1 = 25$ and $\beta_2 = 2.5$.}
\label{fig:reinforcement}
\end{figure*}
\begin{figure}[t!]\centering
\includegraphics[width=0.75\textwidth]{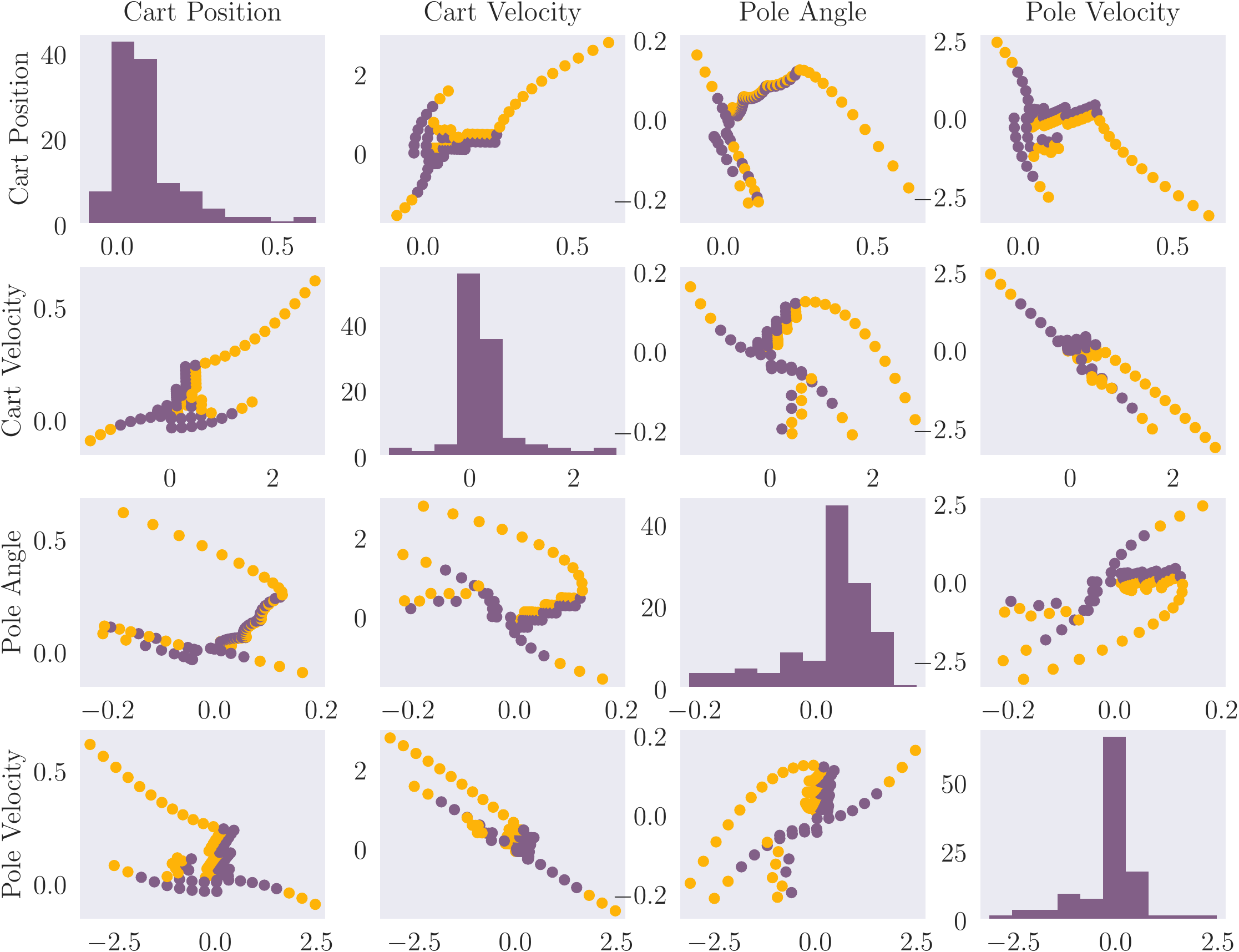}
\caption{Here we show the pairwise state partition learned by the selection policy on the Cart Pole environment. The task is to keep a pole on a moving cart balanced by applying a control signal $a \in \{-1, +1\}$ which moves the cart to the left or the right. On the diagonal we show the histogram (10 bins) of the state dimension over 100 time steps. Note that the matrix is symmetric.}
\label{fig:cartpolepartition}
\end{figure}  
In the following we will show how our hierarchical system is able to solve a continuous reinforcement learning problem using an optimal arrangement of linear control policies. We evaluate on a task known as Acrobot \cite{Sutton1996}, more commonly referred to as the inverted double pendulum. The task is to swing a double-linked pendulum up and keep it balanced as long as possible, where the agent receives a reward of 10 minus a distance penalty between its current state and the goal state. Each episode terminates as soon as the environment reaches a predefined terminal state (hanging downwards) or after 1000 time steps. To balance the pendulum the agent is able to apply a force to the base joint in the range of $a \in [-1, +1]$, thus creating a movement to the left or to the right. This environment poses a non-linear control problem and thus a single linear controller cannot give an optimal solution. We show how using our approach enables a committee of linear experts to efficiently solve this non-linear task. We report the results in Figure~\ref{fig:reinforcement}. We allowed for five experts ($\beta_2 = 2.5$), but our system learns that three experts are sufficient to solve the task. The priors for each expert (lower right Figure, each color represents an expert) center on -1, 0, and 1, which correspond to swinging the double pendulum to the left, no force, and swinging to the right. The remaining two experts overlap accordingly. We can see that the average information processing in the five expert setup decreases, while in the selection it increases to $\log 3$. Both indicate that the system has learned an optimal arrangement of three experts and is thus able to achieve maximum reward and eventually catches up to the performance of a non-linear neural network controller trained with TRPO \cite{Schulman2015} that does not have to struggle with the restriction to linear controllers as our algorithm. Our method successfully learned a partitioning of the double-pendulum state space without having any prior information about any of the system dynamics or the state space. 

In Figure~\ref{fig:cartpolepartition} we show how a model with two linear experts and selector learns to control the cart pole problem. Our method recovers the well known solution in which the pole can be balanced by two linear controllers, where one (dark purple) focuses on keeping the pole upright and the other (dark yellow) on moving the cart such that the other linear controller can take over.

\section{Across Task Specialization for Hierarchical Meta-Learning}

\label{sec:specmeta}
The methods presented in Section \ref{sec:spec} can be easily extended to achieve meta-learning by changing the way the selection mechanism is trained.
Instead of assigning individual states that occur within a task, the selector assigns a complete dataset of a task to an expert. To do so, we must find a feature vector $z(d)$ of the dataset $d$. This feature vector must fulfill the following desiderata: 1) invariance against permutation of data in $d$, 2) high representational capacity, 3) efficient computability, and 4) constant dimensionality regardless of sample size $K$. In the following we propose methods to extract such features for image classification, regression, and reinforcement learning problems -- see Figure \ref{fig:features}.  While the experts are trained on the training dataset $D_\text{meta-train}$, their performance used to optimize the selection policy is based on the validation dataset $D_\text{meta-val}$. The validation dataset contains previously unseen samples that are similar to those in the training set, thus providing a measure for the generalization of each expert. In effect, the selector operates on complete datasets, while the experts operate on single samples.

\subsection{Specialization in Meta-Supervised Learning}
\begin{algorithm}[t]
\caption{Expert Networks for Supervised Meta-Learning.}
\label{alg:meta}
\begin{algorithmic}[1]
\State \textbf{Input}: Data Distribution $p(\altmathcal{D})$, number of samples $K$, batch-size $M$, training episodes $N$\;
\State \textbf{Hyper-parameters}: resource parameters $\beta_1$, $\beta_2$, learning rates $\eta_{x}$, $\eta_x$ for selector and experts\;
\State Initialize parameters $\theta, \vartheta$\;
\For{$i$ = 0, 1, 2, ..., $N$}
\State Sample batch of $M$ datasets $D_i \thicksim p(\altmathcal{D})$, each consisting of a training dataset $D_\text{{meta-train}}$ and a meta-validation dataset\; $D_\text{{meta-val}}$ with $2K$ samples each \;
\For{$D \in D_i$}
\State Find Latent Embedding $z(D_\text{{meta-train}})$ \;
\State Select expert $m \thicksim p_{\theta}(m\vert z(D_\text{{meta-train}})$ \;
\State Compute $\hat{f}(m, D_\text{meta-val})$ \;
\EndFor\
\State Update selection parameters $\theta$ with $\hat{f}(m, D_\text{meta-val})$ \;
\State Update Autoencoder with positive samples in $D_i$ \;
\State Update experts $m$ with assigned $D_\text{{meta-train}}$\;
\EndFor\
\State \textbf{return} $\theta$, $\vartheta$\;
\end{algorithmic}
\end{algorithm}
In a supervised learning task we are usually interested in a dataset consisting of multiple input and output pairs $D = \{(x_i, y_i)\}^N_{i=1}$ and the learner's task is to find a function $f(x)$ that maps from input to output, for example through a deep neural network. To do this, we split the dataset into training and test sets and fit a set of parameters $\theta$ on the training data and evaluate on test data using the learned function $f_\theta(x)$. In meta-learning, we are instead working with meta-datasets $\altmathcal{D}$, each containing regular datasets split into training and test sets. We thus have different sets for meta-training,  meta-validation, and meta-test, i.e.,~ $\altmathcal{D} = \{D_\text{meta-train}, D_\text{meta-val},D_\text{meta-test}\}$. The goal is to train a learning procedure (the meta-learner) that can take as input one of its training sets $D_\text{meta-train}$ and produce a classifier (the learner) that achieves low prediction error on its corresponding test set $D_\text{meta-test}$. The meta-learning is then updated using performance measure based on the learner's performance on $D_{\text{meta-val}}$, compare Algorithm~\ref{alg:meta}. This may not always be the case, but our work (among others, e.g., Finn et al.~ (2017) \cite{Finn2017model}) follow this paradigm. The rationale being that the meta-learner is trained such that it implicitly optimizes the base learner's generalization capabilities. The dataset generating distribution $p(\altmathcal{D})$ is unknown to the learner but remains fixed over course of training. The case where $p(\altmathcal{D})$ is changing is study in the field of (meta) continual learning, but is not the focus of this work.

\begin{figure*}[t!]
\begin{center}
\includegraphics[width=.25\textwidth]{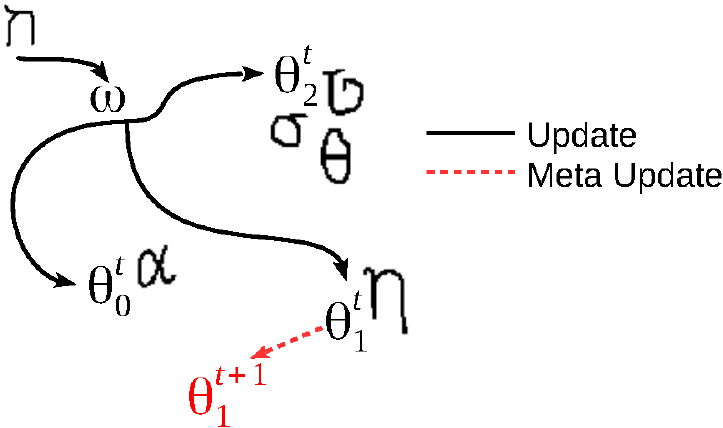}
\hspace{1em}
 \includegraphics[width=.7\textwidth]{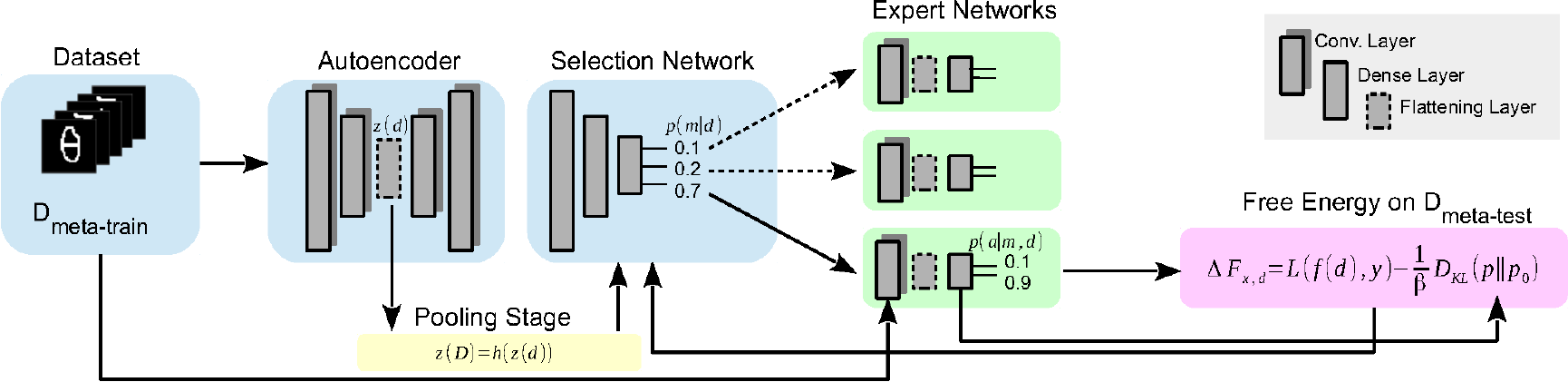}
\end{center}
\caption{\textbf{Left}: The selector assigns the new input encoding to one of the three experts $\theta_0$, $\theta_1$ or $\theta_2$, depending on the similarity of the input to previous inputs seen by the experts. \textbf{Right:} Our proposed method consists of three main stages. First, we feed the training dataset $D_\text{train}$ through a convolutional autoencoder to find a latent representation $z(d_i)$ for each $d_i \in D_\text{train}$, which we get by flattening the preceding convolutional layer (``flattening layer"). We apply a pooling function to the resulting set of image embeddings which serves as input to the selection network.} 
\label{fig:model}
\end{figure*}

For image classification, we propose to pass the images with positive labels in the dataset through a convolutional autoencoder and use the outputs of the bottleneck layer. Convolutional autoencoders are generative models that learn to reconstruct their inputs by minimizing the Mean-Squared-Error between the input and the reconstructed image (see e.g.,\ Vincent et al.\, \cite{vincent2008extracting}. In this way we get similar embeddings $z(d)$ for similar inputs belonging to the same class. We compute the latent representation for each positive sample in $d$ and pass it through a pooling function $h(z(d))$ to find a single embedding for the complete dataset--see Figure~\ref{fig:model} for an overview of our proposed model. We found that max pooling yields the best results, while one could use others, such as mean or min pooling. Yao et al.\ (2019) \cite{yao2019hierarchically} propose a similar feature set.
For regression, we transform the training data into a feature vector $z(d)$ by binning the points into $N$ bins according to their $x$ value and collecting the $y$ value. If more than one point falls into the same bin we average the $y$ values, thus providing invariance against the order of the data in $D_{\text{meta-train}}$. We use this feature vector to assign each dataset to an expert according to $p_\theta(m\vert h(z(D_{\text{meta-train}})))$, which we abbreviate to $p_\theta(m\vert D_\text{meta-train})$. 

In contrast to the objective defined by Equation \eqref{eq:supervisedselectorobj}, the selection policy now selects experts based on their free-energy that is computed over datasets $D_{\text{meta-val}}$ and the selection policy depends on the training datasets $D_{\text{meta-train}}$
\begin{equation}
\max_\theta \E_{p_\theta(m\vert D_{\text{meta-train}})}\left[\hat{f}(m,D_{\text{meta-val}}) - \frac{1}{\beta_1}\log\frac{p_\theta(m\vert D_\text{meta-train})}{p(m)}\right],
\end{equation}
where $\hat{f}(m,D_{\text{meta-val}}) \coloneqq \mathbb{E}_{p_\vartheta(\hat{y}\vert m,x)}\big[-\altmathcal{L}(\hat{y},y) - \frac{1}{\beta_2}\log\frac{p_\vartheta(\hat{y}\vert x,m)}{p(\hat{y}\vert m)}\big]$ is the free energy of expert $m$ on dataset $D_{\text{meta-val}}$, $\altmathcal{L}(\hat{y},y)$ is loss function, and $(x,y) \in D_\text{meta-val}$. The experts optimize their free energy objective on the training dataset $D_{\text{meta-train}}$ defined by
\begin{equation}
\max_\vartheta \E_{p_\vartheta(\hat{y}\vert m,x)}\left[-\altmathcal{L}(\hat{y}, y) - \frac{1}{\beta_2}\log\frac{p_\vartheta(\hat{y}\vert x,m)}{p(\hat{y}\vert m)}\right],
\end{equation} 
where $(x, y) \in D_\text{meta-train}$.

\subsection{Specialization in Meta-Reinforcement Learning}
\begin{algorithm}[t]
\caption{Expert Networks for Meta-Reinforcement Learning.}
\label{alg:metarl}
\begin{algorithmic}[1]
\State \textbf{Input}: Environment Distributions $p(T)$ and $p(T')$, number of roll-outs $K$, batch-size $M$, training episodes $N$, number of tuples $L$ used for expert selection\;
\State \textbf{Hyper-parameters}: resource parameters $\beta_1$, $\beta_2$, learning rates $\eta_{x}$, $\eta_x$ for selector and experts\;
\State Initialize parameters $\theta, \vartheta$\;
\For{$i$ = 1, 2, 3, ..., $N$}
\State Sample batch of $M$ environments $E_i^{\text{meta-train}} \thicksim p(T)$\;
\For{$E \in E_i^{\text{meta-train}}$}
\For{$k$ = 1, 2, 3, ..., $K$}
\State Collect $\tau = \{(x_t, a_t, r_t, t)\}_{t=1}^L$ tuples by following random expert \;
\State Select expert $m \thicksim p_{\theta}(m\vert\tau)$ with RNN policy \;
\State Collect trajectory $\tau_k = \{(x_t, a_t, r_t, t)\}_{t=L}^T$ by following $p_\vartheta(a|x,m)$
\EndFor\
\State Compute $F_t = \sum_{l=0}^T\gamma^l f(x_{t+l}, m_{t+l}, a_{t+l})$ for trajectories $\tau$\;
\State where $f(x,m,a) = r_{\text{meta-train}}(x, a) - \frac{1}{\beta_2} \log\frac{p_{\vartheta}(a\vert x,m)}{p(a\vert m)}$.
\EndFor\
\State Compute $\bar{F}_t = \sum_{l=0}^T\gamma^l \bar{f}(x_{t+l}, m_{t+l})$ with \;
\State $\bar{f}(x,m) = \mathbb{E}_{p_\vartheta(a\vert x,m)}\left[r_{\text{meta-train}}(x,a) - \frac{1}{\beta_2} \log \frac{p(a\vert x, m)}{p(a\vert m)}\right]$ \;
\State Update selection parameters $\theta$ with $\bar{F}$ collected in batch $i$ \;
\State Update experts $m$ with roll-outs collected in batch $i$\;
\EndFor\
\State \textbf{return} $\theta$, $\vartheta$\;
\end{algorithmic}
\end{algorithm}

In meta reinforcement learning  we extend the problem to a set of tasks $t_i \in T$, where a MDP $t_i = (\altmathcal{S}, \altmathcal{A}, P_i, r_i)$ defines each task $t_i$. We are now interested in finding a set of policies $\Theta$ that maximizes the average  cumulative reward across all tasks in $T$ and generalizes well to new tasks sampled from a different set of tasks $T'$.

In this setting we use a dynamic recurrent neural network (RNN) with independent recurrent units \cite{li2018independently} to classify trajectories and a second RNN to learn the value function (see Appendix \ref{app:experiments} for details). We feed the RNN with $L$ $(s_t, a_t, r_t, s_{t+1})$ tuples to describe the underlying Markov Decision Process describing the task. At $t = 0$ we sample the expert  according to the learned prior distribution $p(m)$, as there is no other information available until we have collected $L$ samples at which point an expert is selected -- see Algorithm \ref{alg:metarl}. Lan et al.\ (2019) \cite{lan2019meta} propose a similar feature set. Importantly, the expert policies are trained on the meta-training environments, but evaluated on unseen but similar validation environments. In this setting we define the discounted free energy $\bar{F}_t$ for the selector as
\begin{equation}
\bar{F}_t = \sum_{l=0}^T\gamma^l \bar{f}(x_{t+l}, m_{t+l}),
\end{equation}
with $\bar{f}(x,m) = \mathbb{E}_{p_\vartheta(a\vert x,m)}\left[r_{\text{meta-train}}(x,a) - \frac{1}{\beta_2} \log \frac{p(a\vert x, m)}{p(a\vert m)}\right]$, where $r_{\text{meta-val}}$ is a reward function defined by a validation environment (see Figure \ref{fig:metarl} for details).

\section{Experimental Results: Across Task Specialization and Meta-Learning}
\label{sec:experimentsmeta}
In this section we present our experimental results in the meta-learning domain. To show the flexibility of our method we evaluate on regression, classification and reinforcement learning problems. In regression, we evaluate how our method adapts to changing sine functions, for classification we look at the Omniglot dataset \cite{lake2015human}. To evaluate on reinforcement learning we introduce a range of versions of the double pendulum task \cite{Sutton1996}. We provide all experimental details such as network architectures and hyper-parameters in Appendix \ref{app:meta}.
\begin{figure*}[t!]
\centering
    \begin{minipage}{.33\textwidth}
    \centering
    \includegraphics[width=\linewidth]{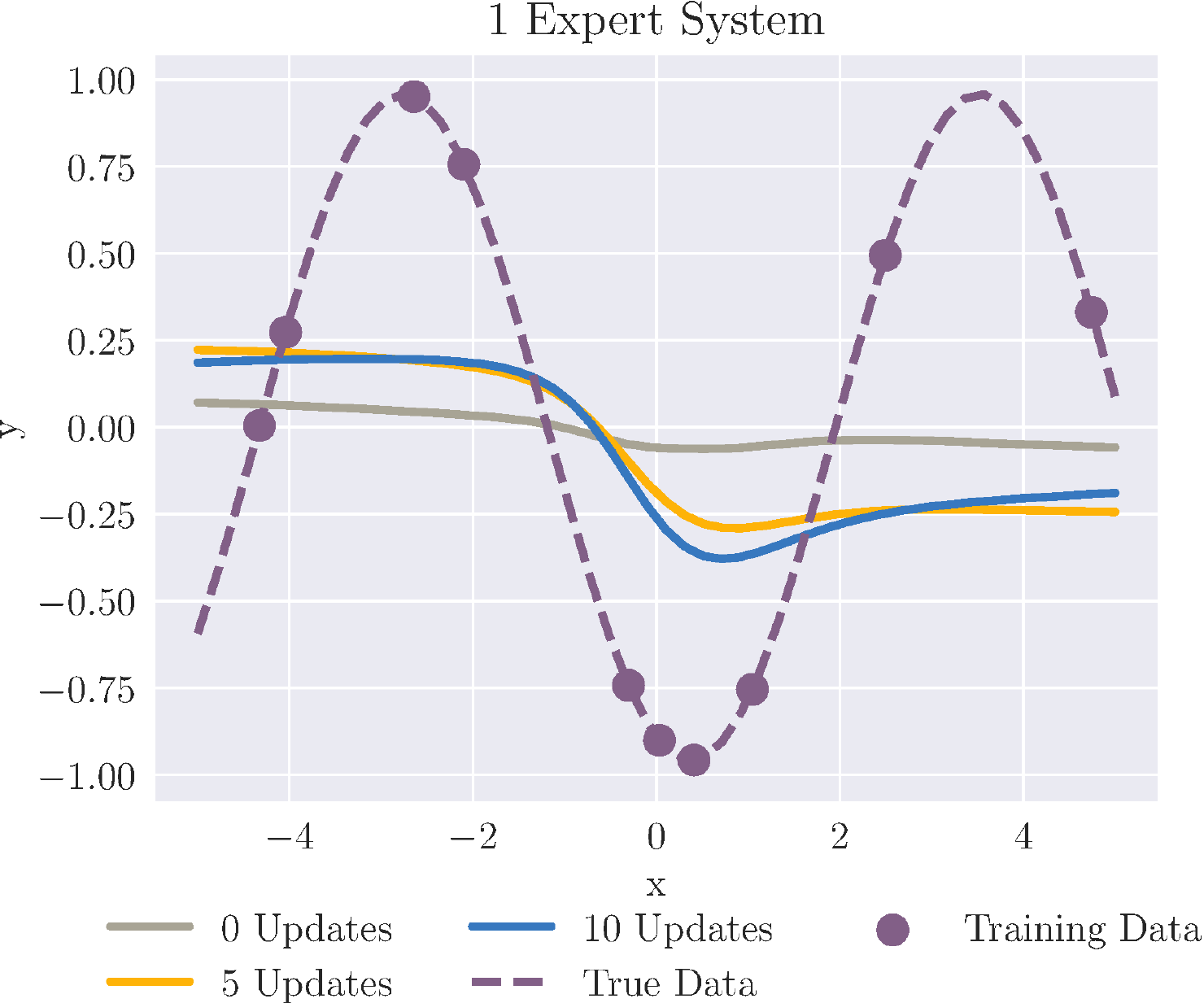}
  \end{minipage}
  \begin{minipage}{.33\textwidth}
  \centering
    \includegraphics[width=\linewidth]{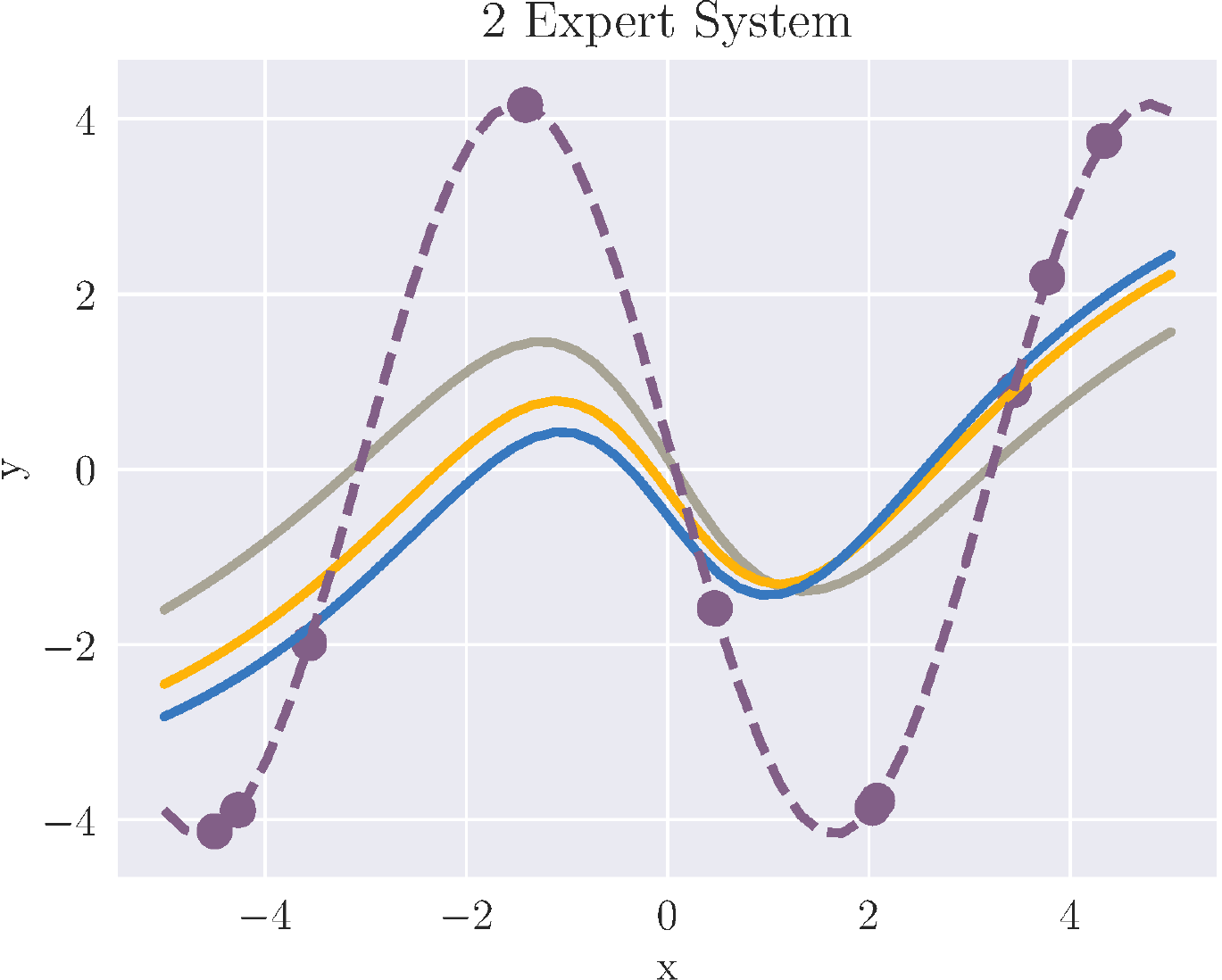}
  \end{minipage}
    \begin{minipage}{.33\textwidth}
    \centering
    \includegraphics[width=\linewidth]{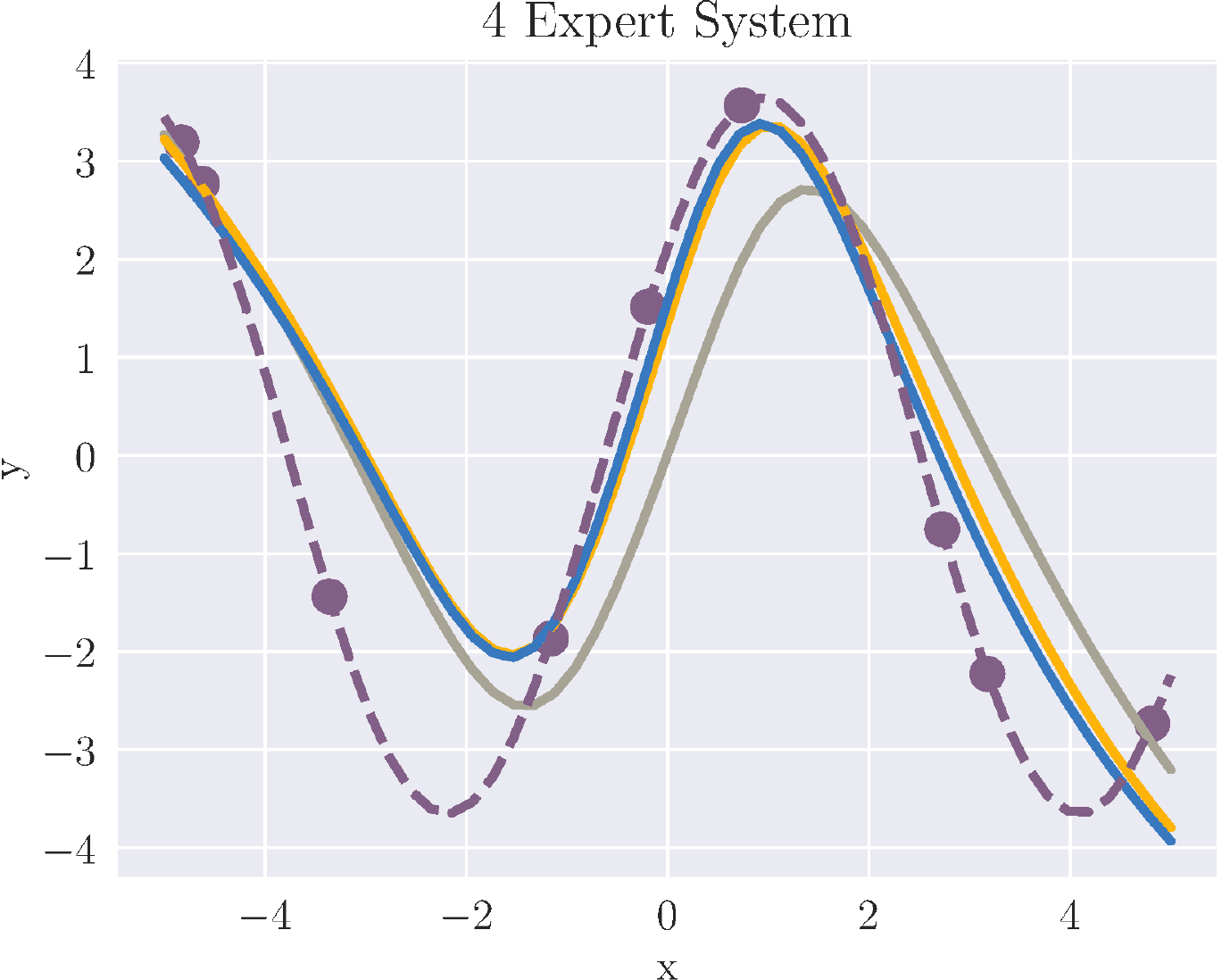}
  \end{minipage}
    \begin{minipage}{.33\textwidth}
    \centering
    \includegraphics[width=\linewidth]{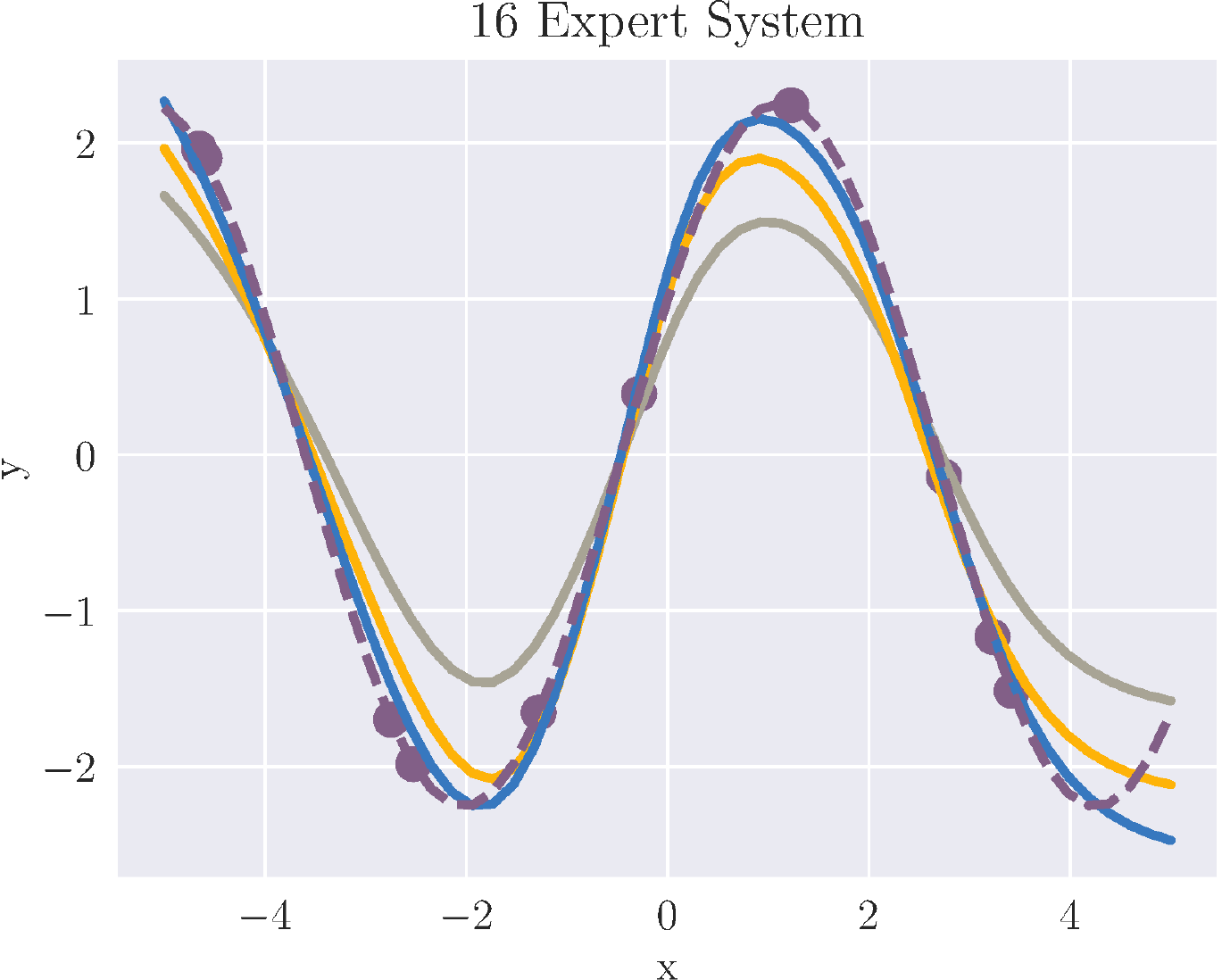}
  \end{minipage}
  \begin{minipage}{.67\textwidth}
  \centering
  \includegraphics[width=\textwidth]{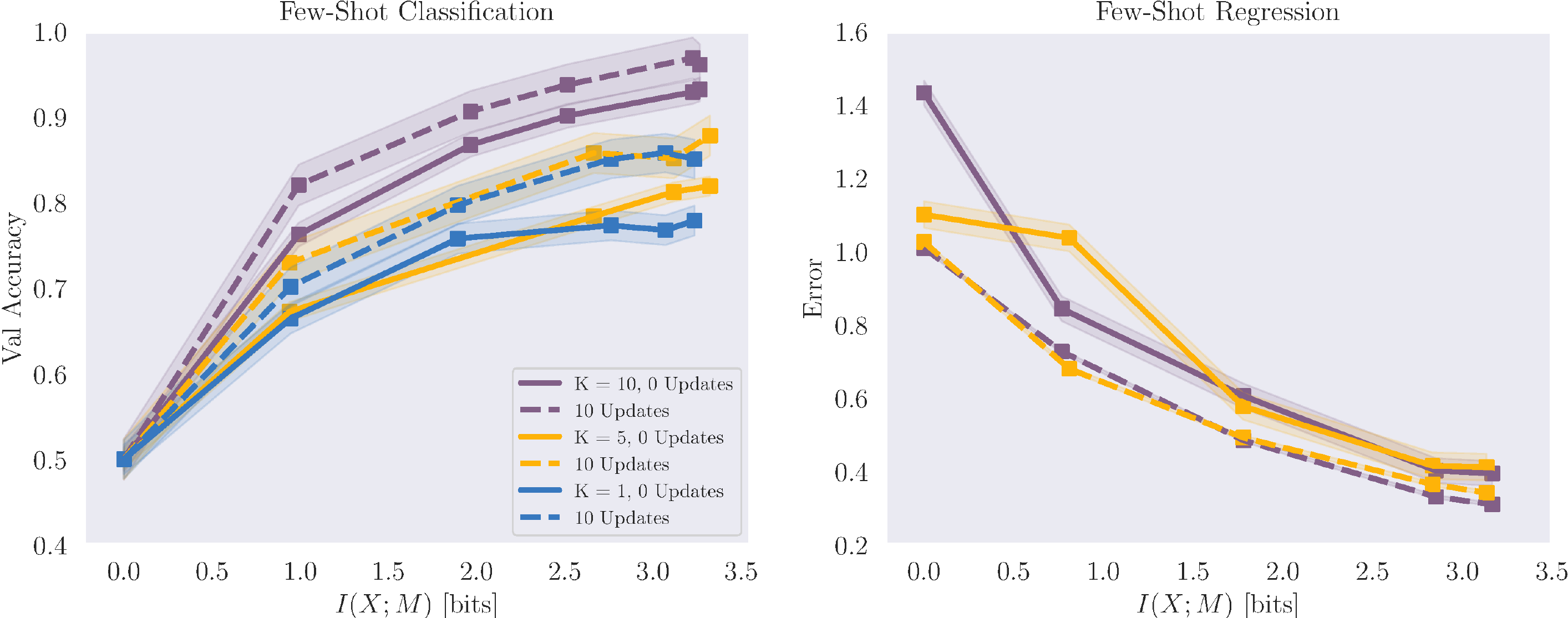}
  \end{minipage}
\caption{Here show how the system is able to adapt to new problems as the number of experts increases. The single expert system is not able to learn the underlying structure of the sine wave, where the two expert system is already able to capture the periodic structure. Adding more experts improves adaptation further, as the results show. Each expert is a shallow neural network with a single hidden layer and and output layer (see Appendix \ref{app:experiments} for details). In the bottom row we show the rate-utility curve describing the trade-off between information processing and expected utility (transparent area represents one standard deviation), where increasing $I(X;M)$ improves adaptation. To obtain these results we set $\beta_1 = 25$ and $\beta_2 = 1.25$.} 
\label{fig:sine}
\end{figure*}
\subsection{Sinusoid Regression}
\label{sec:experimentssine}

We adopt this task from Finn et al.~ (2017) \ \cite{Finn2017model}. In this $K$-shot problem, each task consists of learning to predict a function of the form $y = a \cdot \sin(x + b)$, with both $a \in [0.1, 5]$ and $b \in [0, 2\pi]$ chosen uniformly, and the goal of the learner is to find $y$ given $x$ based on only $K$ pairs of $(x, y)$. Given that the underlying function changes in each iteration it is impossible to solve this problem with a single learner. As a loss function we use Mean-Squared-Error and the dataset embedding is described in Algorithm \ref{alg:meta}. Each expert is a shallow neural network consisting of a single hidden layer connected to an output layer (see Appendix \ref{app:experiments} for details).
Our results show that by combing expert networks, we are able to reduce the generalization error iteratively as we add more experts to our system--see Figure~\ref{fig:sine} for $K = 5$ and $K = 10$ settings. In Figure~\ref{fig:sine} we show how the system is able to capture the underlying problem structure as we add more experts  and in Figure~\ref{fig:sinepartition} we visualize how the selector's partition of the problem space looks like. In Appendix \ref{app:experiments}, Figure~\ref{fig:sineapp} we show additional results and give an overview of our algorithm in Algorithm \ref{alg:meta}.

\subsection{Few-Shot Classification}
\label{sec:experimentsfew}
A special case of meta-learning for classification are $K$-Shot $N$-way tasks, where a training set consists of $K$ labeled examples of each of the $N$ classes ($K\cdot N$ examples per dataset) and a corresponding test set is also provided. In our study, we focus on the following variation of $K$-Shot $N$-Way tasks: $2K$ samples ($K$ positive and $K$ negative examples) define a training dataset which the meta-learner must assign to an expert learner that has a specialized policy to classify the $2K$ samples. To evaluate experts performance we use a meta-validation set consisting of a different set of $2K$ samples. Note that we draw the negative examples from any of the remaining classes.

The Omniglot dataset \cite{lake2011one} consists of over 1600 characters from 50 alphabets (see Figure~\ref{fig:alphabetassignment} for examples ). As each character has merely 20 samples each drawn by a different person, this forms a difficult learning task and is thus often referred to as the ''transposed MNIST'' dataset. The Omniglot dataset is a standard meta-learning benchmarking dataset \cite{Finn2017model,vinyals2016matching,ravi2017optimization}.	
\begin{table}[t]
\centering
\begin{tabular}{*6c}
\toprule
\multicolumn{6}{c}{\textsc{\normalsize Omniglot Few-Shot Results}}  \\
\toprule
{} & \multicolumn{3}{c}{\textsc{One Conv. Block Baselines}} &  \multicolumn{2}{c}{\textsc{Methods}}  \\
\midrule
{} & \textsc{Pre-Training} & \textsc{MAML} &  \textsc{Matching Nets} & \textsc{MAML} & \textsc{Matching Nets}  \\
\midrule
\textbf{K} & \textbf{\% Acc} &  \textbf{\% Acc} &  \textbf{\% Acc} &\textbf{\% Acc} & \textbf{\% Acc} \\
\midrule
\textbf{1} & 50.6 ($\pm$ 0.03) & 81.2 ($\pm$ 0.03) &  52.7 ($\pm$ 0.05) & 95.2 ($\pm$ 0.03) &  95.0 ($\pm$ 0.01)\\
\textbf{5} & 54.1 ($\pm$ 0.09) & \textbf{88.0} ($\pm$ 0.01) & 55.3 ($\pm$ 0.04) & 99.0 ($\pm$ 0.01) &  98.7 ($\pm$ 0.01)\\
\textbf{10} & 55.8 ($\pm$ 0.02) & 89.2 ($\pm$ 0.01) & 60.9 ($\pm$ 0.06) & 99.2 ($\pm$ 0.01) &  99.4 ($\pm$ 0.01) \\
\bottomrule
\multicolumn{5}{c}{\textsc{Our Method}} \\
\midrule
\multicolumn{5}{c}{\textsc{Number of Experts}} \\
\midrule
{} &  \multicolumn{2}{c}{\textbf{2}} & \multicolumn{2}{c}{\textbf{4}}\\
\midrule
{} & \textbf{\% Acc} & \textbf{I(M;X)} & \textbf{\% Acc} & \textbf{I(M;X)} \\
\midrule
\textbf{1} & 66.4 ($\pm$ 0.02) & 0.99 ($\pm$ 0.01) & 75.8 ($\pm$ 0.02) & 1.96 ($\pm$ 0.01)\\
\textbf{5} & 67.3 ($\pm$ 0.01) & 0.93 ($\pm$ 0.01) & 75.5 ($\pm$ 0.01) & 1.95 ($\pm$ 0.10) \\
\textbf{10} & 76.2 ($\pm$ 0.04)  & 0.95 ($\pm$ 0.30) & 86.7 ($\pm$ 0.01) & 1.90 ($\pm$ 0.03) \\
\midrule
{} & \multicolumn{2}{c}{\textbf{8}} & \multicolumn{2}{c}{\textbf{16}} \\
\midrule
{} & \textbf{\% Acc} & \textbf{I(M;X)} & \textbf{\% Acc} & \textbf{I(M;X)} \\
\midrule
\textbf{1} & 77.3 ($\pm$ 0.01) & 2.5 ($\pm$ 0.02) & \textbf{82.8} ($\pm$ 0.01) & 3.2 ($\pm$ 0.03) \\
\textbf{5} & 78.4 ($\pm$ 0.01) & 2.7 ($\pm$ 0.01) & 85.2 ($\pm$ 0.01) & 3.3 ($\pm$ 0.02) \\
\textbf{10} & 90.1 ($\pm$ 0.01) & 2.8 ($\pm$ 0.02) & \textbf{95.9} ($\pm$ 0.01) & 3.1 ($\pm$ 0.02)\\
\bottomrule
\end{tabular}
\caption{Classification accuracy after 10 gradient steps on the validation data. Adding experts consistently improves performance, obtaining the best results with an ensemble of 16 experts. Pre-training refers to a single expert system trained on the complete dataset. Our method outperforms the pre-training, Matching Nets, and the MAML baseline (see Appendix \ref{app:experiments} for experimental details), when the network architecture is reduced to a single convolution block. This corresponds to our expert network architecture. Using the suggested architectures by the respective studies, we achieve classification accuracy $\geq 95\%$. In this experiment we set $\beta_1 = 20.0$ and $\beta_2 = 2.5$ for 2 and 4 experts and $\beta_1 = 50.0$ and $\beta_2 = 1.25$ for 8 and 16 experts.}
\label{tab:classtable}
\end{table}

\begin{figure*}[t!]
\begin{minipage}{\textwidth}
\centering
\begin{tabular}{lcc}
\toprule
& \multicolumn{2}{c}{\textbf{Task Distribution}}\\
\textbf{Paramater} & \textbf{$T$} & \textbf{$T'$}  \\
\midrule
Distance Penalty & [$10^{-3},10^{-1}$] & [$10^{-3},10^{-2}$] \\
Goal Position & [0.3, 0.4] & [0, 3]\\
Start Position & [-0.15, 0.15] & [-0.25, 0.25] \\
Motor Torques & $[0, 5]$ & $[0, 3]$\\
Motor Actuation & [185, 215] & [175, 225] \\
Inverted Control & $p = 0.5$ & $p = 0.5$  \\
Gravity & [0.01, 4.9] & [4.9, 9.8] \\
\bottomrule
\end{tabular}
\end{minipage}
\begin{minipage}{\textwidth}
\centering
\includegraphics[width=0.85\textwidth, trim={5.75cm 0cm 29cm 1cm}, clip]{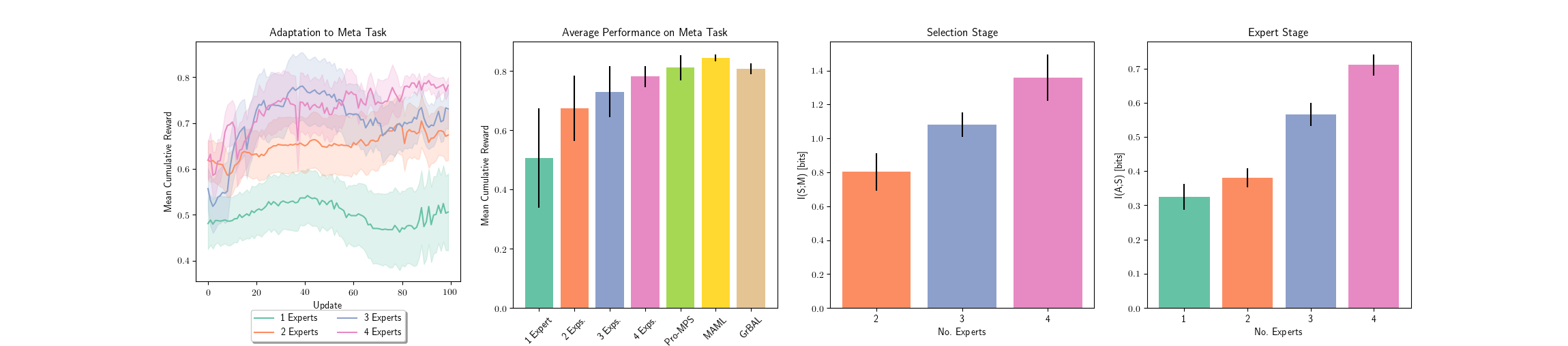}
\includegraphics[width=0.85\textwidth, trim={29.5cm 0.15cm 5cm 1cm}, clip]{meta_rl_new.png}
\end{minipage}
\caption{We sample all parameters uniformly from the specified range for each environment, where we use $T$ for training and $T'$ for meta evaluation. During training  we draw environments from $T$, but evaluate on a different environment also drawn from $T$ to measure generalization. The agent achieves higher reward when adding more experts while the information-processing of the selection and of the expert stage increases, indicating that the added experts specialize successfully. We achieve comparable results to MAML \cite{Finn2017model}, Proximal Meta-Policy Search (Pro-MPS) \cite{rothfuss2018promp}, and GrBAL \cite{nagabandi2018learning}. Shaded areas and error bars represent one standard deviation. See Appendix \ref{app:experiments} for experimental details.}
\label{fig:metarl}
\end{figure*}

We consider three experimental setups: 1) how does a learner with only a single hidden layer perform when trained na\"{i}vely compared to with sophisticated methods such as MAML \cite{Finn2017model} and Matching Nets \cite{vinyals2016matching} as a baseline? 2) does the system benefit from adding more experts and if so, at what rate? and 3) how does our method compare to the aforementioned algorithms? Regarding 1) we note that introducing constraints by reducing the representational power of the models does not facilitate specialization is it would by explicit information-processing constraints. In the bottom row of Figure \ref{fig:sine} we address question 2).  We can interpret this curve as the rate-utility curve showing the trade-off between information processing and expected utility (transparent area represents one standard deviation), where increasing $I(X;M)$ improves adaptation. The improvement gain grows logarithmically, which is consistent with what rate-distortion theory would suggest. In Table \ref{tab:classtable} we present empirical results addressing question 3).

We train the learner on a subset of the dataset ($\approx 80\%$, $\approx$ 1300 classes) and evaluate on the remaining $\approx 300$ classes, thus investigating the ability to generalize to unseen data. In each round we build the datasets $D_{\text{meta-train}}$ and $D_{\text{meta-test}}$ by selecting a target class $c_t$ and sample $K$ positive and $K$ negative samples. To generate negative samples we draw $K$ images randomly out of the remaining $N-1$ classes. We present the selection network with the feature representation of the $K$ positive training samples (see Figure~\ref{fig:model}), but evaluate the experts' performance on the $2K$ test samples in $D_{\text{meta-test}}$. We can interpret the free energy of the experts in this setting as a measure of how well the expert is able to generalize to new samples of the target class. Using this optimization scheme, we train the expert networks to become \textit{experts} in recognizing a subset of classes. We train the experts using the $2K$ samples from the training dataset that the selection network assigned to an expert---see Table \ref{tab:classtable} for results. We followed the design of Vinyals et al.\, \cite{vinyals2016matching} to design our experts but reduce the number of blocks to one to introduce effects of resource limitation, whereas in the original study the authors used four blocks. The single convolutional block consists of 32 3$\times$3 filters with strided convolutions followed by a batch normalization layer and a ReLu non-linearity. The output is fed into a softmax layer giving a distribution over the classes (see also Appendix \ref{app:experiments}). This reduced representational capacity drives specialization, as the expert can not reliably classify all characters from the training data, but a subset is feasible (see also Figure~\ref{fig:alphabetassignment}). To evaluate our method we compare different ensemble sizes against three baselines: pre-training, MAML \cite{Finn2017model} and Matching Networks \cite{vinyals2016matching}. In the pre-training setting we train a single convolutional neural network on batches drawn from the training data  and evaluate on validation data by allowing 10 gradient steps for fine-tuning. Note, that when using the architecture of 4 blocks as suggested in the original paper \cite{vinyals2016matching,Finn2017model}, we are able to achieve $\geq$95\% accuracy on the test data in both MAML and matching nets, but not on the pre-training setting.

\subsection{Meta Reinforcement Learning}
\label{sec:experimentsmetarl}
In meta reinforcement learning the goal is to find a policy that performs well over a set of tasks. 
We create a set of RL tasks by extending the ``Inverted Double Pendulum problem" \cite{Sutton1996} implemented in OpenAI Gym \cite{Brockman2016} by allowing a broad range of task parameters. Each time we create a new environment we sample from a specified distribution, where we modify inertia, motor torques, reward function, goal position and invert the control signal -- see Table \ref{fig:metarl} for details. We create one environment per training episode, where during a single training episode parameters remain unchanged. We measure the free energy of an expert on a second task with parameters also drawn from $T$. 

To evaluate the agents' meta-learning capabilities we define a second set of tasks $T'$ where the parameter distributions are different, providing new but similar reinforcement learning problems. In each episode we sample $M$ environments from $T$ and update the system batch-wise. After training we evaluate on tasks from $T'$, thus testing the agents generalization. We trained the  system for 1000 Episodes with 64 tasks from $T$ and evaluate for 100 system updates on tasks from $T'$. We report the results in Figure~\ref{fig:metarl}, where we can see improving performance as we add more experts, where the mutual information characterizing the selection stage indicates that the selector is able to identify suitable experts for the tasks.

\begin{figure*}[t!]
\centering
\includegraphics[width=0.75\textwidth]{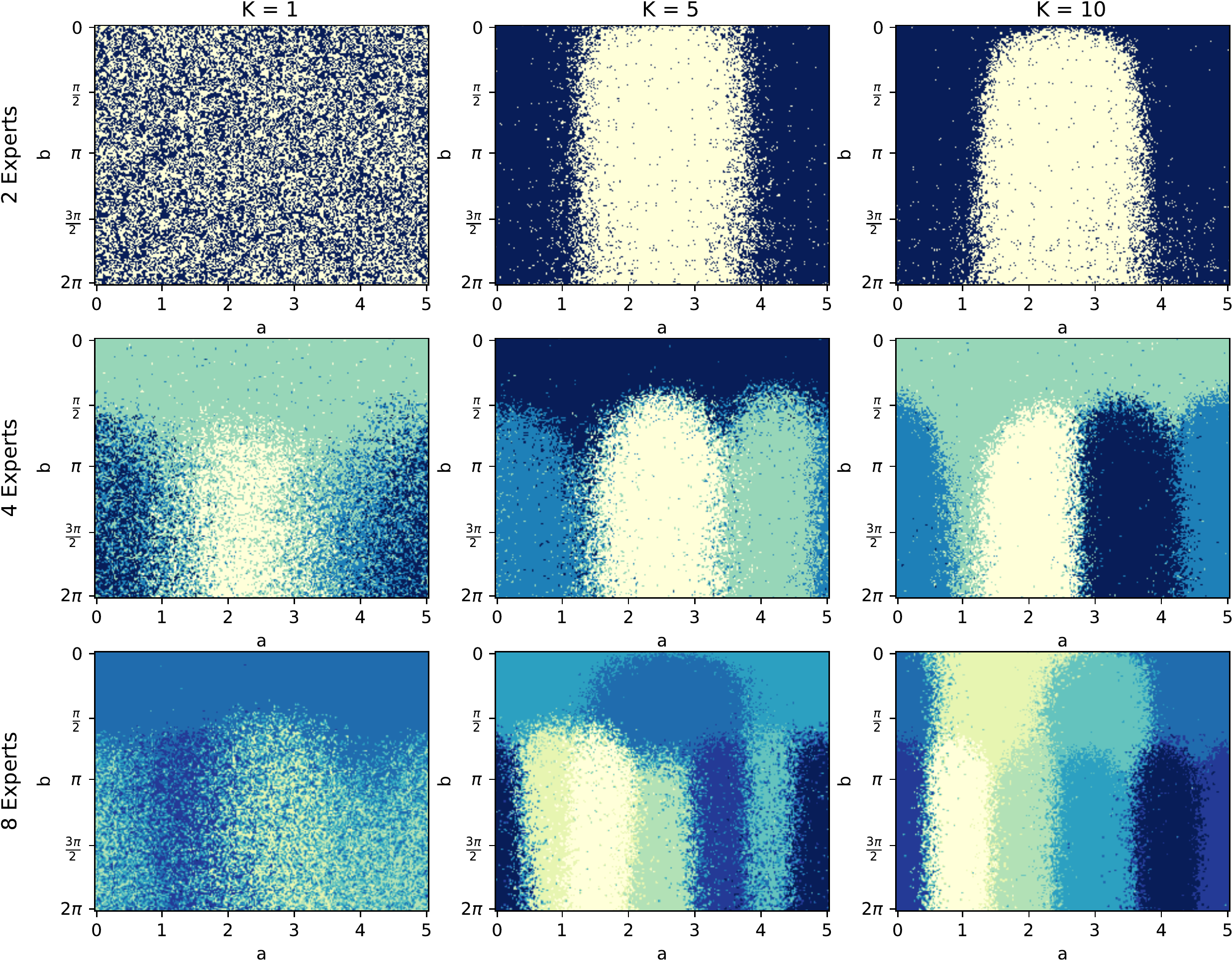}
\caption{Here we show the soft-partition found by the selection policy for the sine prediction problem $y = a\cdot\sin(x + b)$, where we sample $a,b$ uniformly at each trial and each color represents an expert. To generate these plots we train a system on $K=1, 5$ or $10$, sample $a,b$ and $K$ points and feed the dataset to the selection policy. We can see that the selection policy becomes increasingly more precise as we provide more points per dataset (denoted by $K$) to the system.}
\label{fig:sinepartition}
\end{figure*}

\section{Discussion}
\label{sec:disc}
\subsection{Analyzing the Discovered Meta-Knowledge}
\label{sec:analyzingmeta}
\begin{figure}
\includegraphics[width=0.85\textwidth]{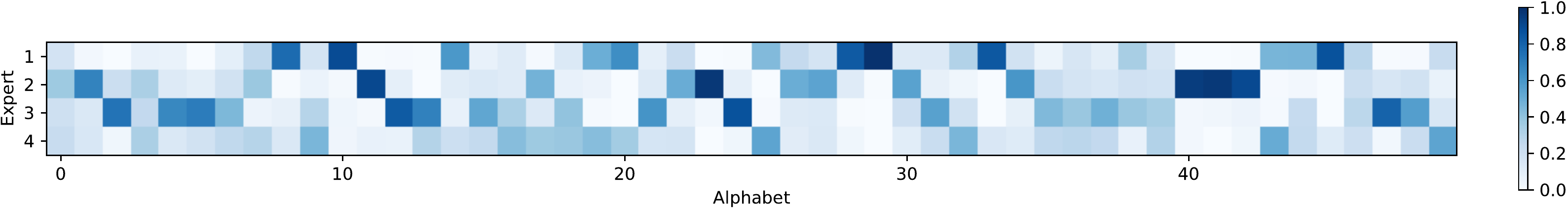}
\centering
\includegraphics[width=0.4\textwidth]{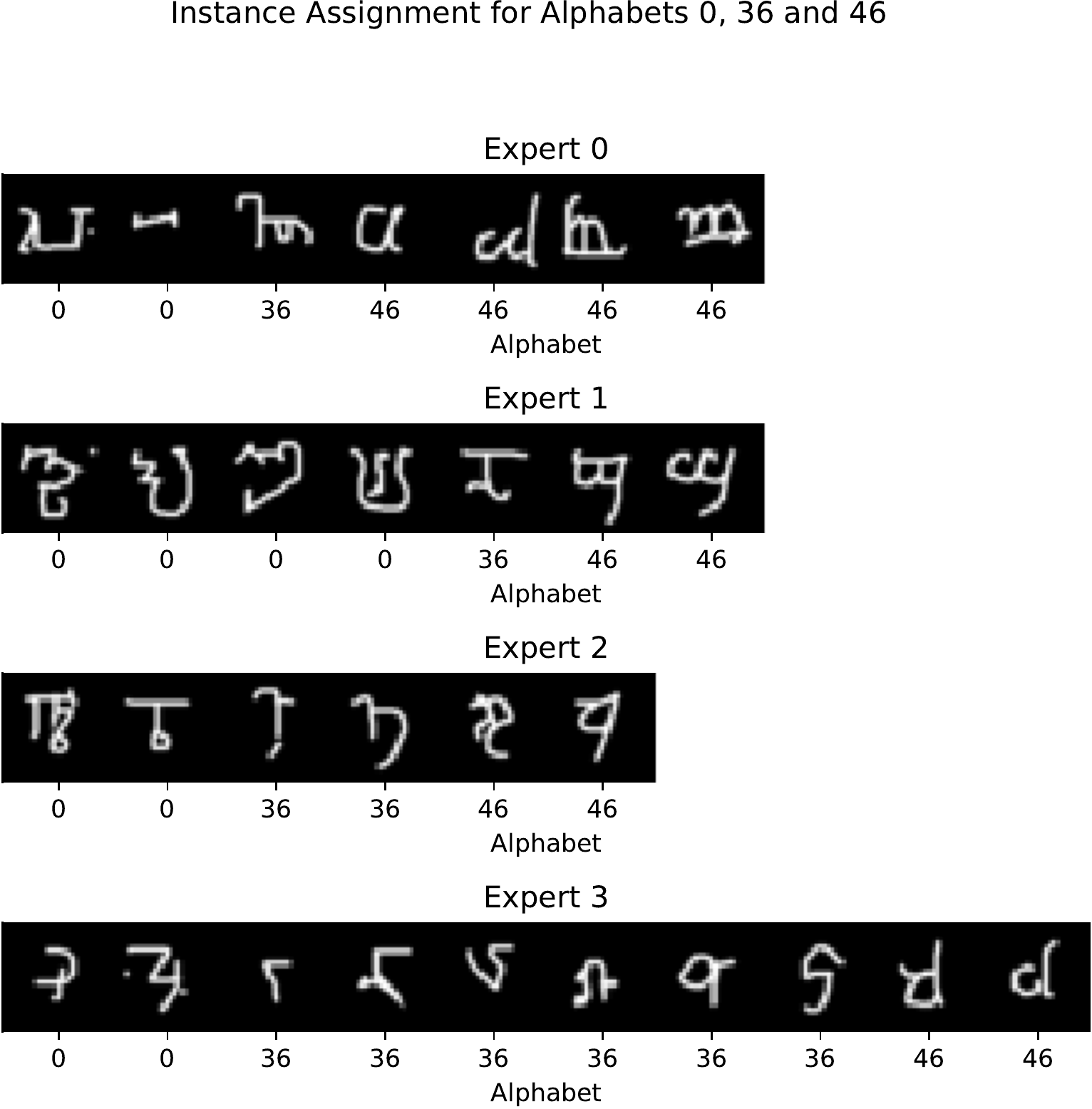}
\includegraphics[width=0.4\textwidth]{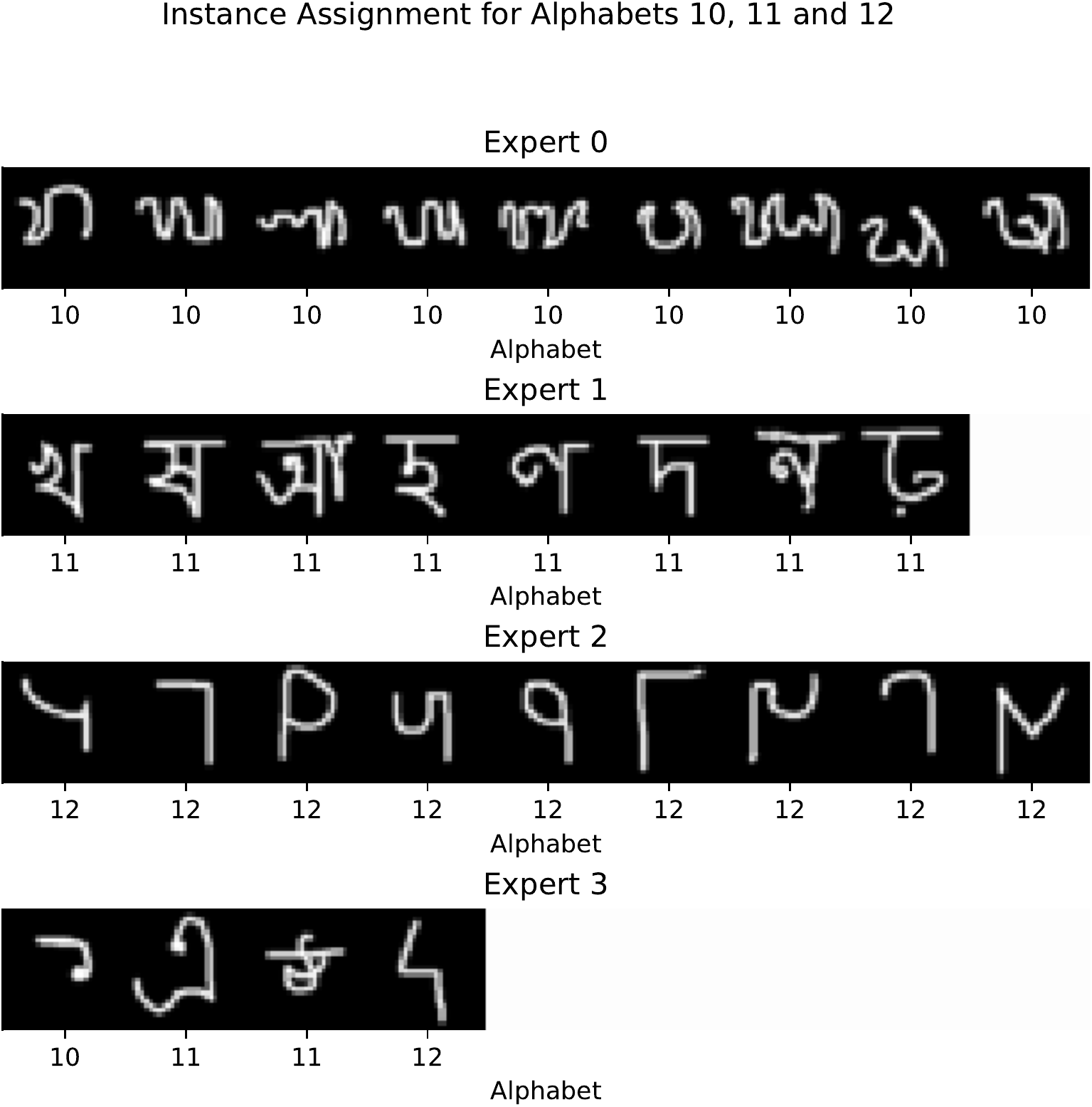}
\caption{\textbf{Upper Row:} The system learns to assign characters from the same alphabet to the same expert. This happens without any prior information of the concept of alphabets or any other label provided. \textbf{Lower Row:} Here we report selection results on 15 characters sampled from 3 alphabets. To illustrate how the system operates we first show characters from alphabets that the system has difficulties assigning to experts (here alphabets 0, 36, and 46) in the lower left figure. In the lower right figure we show characters from alphabets that the selector assigns with high confidence (10, 11, and 12). }
\label{fig:alphabetassignment}
\end{figure}
In contrast to monolithic approaches that train a single agent as a black-box we can analyze and interpret the meta-knowledge discovered by our hierarchical method. In the following we will discuss this in the supervised and reinforcement learning setting.

In meta-regression the problem space defined by a set of sine functions $y = a\cdot \sin (x+b)$ is split among the ensemble of expert regressors based on only $K \in \{1, 5, 10\}$. As expected, the assignment becomes  more accurate the more samples we have -- see Figure~\ref{fig:sinepartition} where we report how the selection network partitions the sine task space. The shape of the emerged clusters indicates that the selection is mainly based on the amplitude $a$ of the current sine function, indicating that from an adaptation point-of-view it is more efficient to group sine functions based on amplitude $a$ instead of phase $b$. We can also see that an expert specializes on the low values for $b$ as it covers the upper region of the $a\times b$ space. The selection network splits this region among multiple experts if we increase the set of experts to 8 or more.

We can also analyze the class assignment policy learned by the selector network. The assignment map in Figure~\ref{fig:alphabetassignment} suggests that the selector learns to group characters by their alphabet based only on features. The selector's policy spreads different characters from the same alphabet (e.g.,~alphabets 0, 36, and 46) across multiple experts while assigning similar characters from different alphabets to the same experts. This specializations gives rise to the meta-learning ability as it is able to adapt expert parameters within only a few gradient steps. We generated this plot by passing images of characters through the system (i.e.,~computing their latent embedding and assigning them to experts) after training is complete to obtain an overview of the class distribution.

For reinforcement learning we demonstrate this idea by looking at how the selection networks splits the state space into to linearly controllable regions in Figure~\ref{fig:cartpolepartition}. We can see that one expert receives the states around zero (dark purple) and the other experts sees only  states near the boundary (dark yellow). We derive from this that the purple expert specializes on balancing the pole and the yellow expert specializes on moving the cart such that the other expert can easily balance the pole. This is consistent with the fact that a linear policy can balance the pole when it is close to standing upwards.

\subsection{Critical issues}
\label{sec:critical}
A limitation of our method is low sample efficiency in the RL domain. To alleviate this problem one could imagine to combine our system with model-based RL methods which promise to reduce the number of samples the agent needs to see in order to learn efficiently. Another research direction would be to investigate our systems performance in continual adaptation tasks, such as in the work of Yao et al.\ (2019) \cite{yao2019hierarchically}. There the agent is continuously exposed to datasets (e.g.,\ additional classes and samples). The restriction to binary meta classification tasks is another limitation of our method, which we leave for feature work. 

Another open question remains the tuning of $\beta$ values. As the utility function can in principle be unbounded whereas the information-processing quantities are obviously bounded, the agent may be faced with learning two values that differ greatly in magnitude. This becomes especially challenging in reinforcement learning scenarios, where a value function of varying magnitude has to be learned. This poses a difficult learning problem and there have been several proposals to tackle this. A method dubbed ''Pop-Art'' proposed by van Hasselt et al.~ (2016) \cite{vanHasselt2016learning}, where they treat the incoming utility values as a stream of data and normalize the values to given range. In the reinforcement learning setup we also tried cooling schedule for $\beta$, as suggested by Fox et al.~ (2015) \cite{fox2016taming}. In their work the authors propose to change the weight of the entropy penalty in MaxEnt RL as time progresses, thus encouraging exploration in the beginning and penalizing it the more information an agent has gathered. We did not observe any significant performance gains. 

The specific value of $\beta$ depends on the scale of the utility function. As the value of $\beta$ strongly influences the outcome of the experiments, it must be chosen with care and comes with the same limitations as any other regularization technique. If it is chosen to small, the regularization term dominates the utility term and the experts are not able to learn anything. On the other hand, if it is set to a large value, the regularization term vanishes, such that no specialization can occur and thus the selector has a hard time assigning data to experts. To remedy this, there are in principle two ways to choose $\beta$: one is to set a information-processing limit for each expert and then (manually or with some optimization technique) tune beta such hat this constraint is satisfied. This has the advantage that this value can be interpreted, e.g., "the experts can process 1 bit of information, i.e.,~ distinguishing two options". The other way is to run a grid search over a pre-defined range and chose the one that fits best. In this work, we used the second strategy.

\subsection{Related Work}
\label{sec:relatedwork}
Investigating information-theoretic cost functions and constraints in learning systems has recently enjoyed increasing interest in machine learning \cite{grau2018soft,galashov2019information,grover2019uncertainty,tschannen2019mutual}. The method we propose in this study falls into a wider class of algorithms that aim to deal more efficiently with learning and decision-making problems~\cite{Daniel2012,Neumann2013,Martius2013,Leibfried2015,Grau-Moya2016,Peng2017,Grau-Moya2017,Ghosh2017,Hihn2018,Schach2018,Gottwald2019,gottwald2019bounded}. 

Applying such constraints to methods for reinforcement learning is often motivated by the aim of stabilizing learning and reducing sample complexity. One such approach is Trust Region Policy Optimization (TRPO) introduced by Schulman et al.\ (2015) \cite{Schulman2015}, where a $\DKL$ penalty between the old and the new policy is imposed to limit update steps, providing a theoretical guarantee for monotonic policy improvement. In our approach we define this region by $\DKL$ between the agent's posterior and prior policy, thus allowing to learn this region and to adapt it over time.  This basic idea has been extend to meta-learning by \cite{rothfuss2018promp}, which we use to compare our method against in meta-rl experiments. Daniel et al.\ (2012) follow a similar idea with relative entropy policy search methods \cite{Daniel2012}. The algorithm builds on learning a gating policy that can decide which sub-policy to choose. The authors impose a $\DKL$ constraint between the data distribution and the next policy to achieve this. Both these approaches smoothen the updates as large steps are discouraged. Our method follows the same principle, but we enforce small updates by discouraging deviation from the agent's prior and by limiting the representational power (e.g.,~by linear decision makers). 

The hierarchical structure we employ is related to Mixture of Experts (MoE) models. Jacobs et al.\ (1991) \cite{Jacobs1991} introduced MoE as tree structured models for complex classification and regression problems, where the underlying approach is the divide and conquer paradigm. As in our approach, three main building blocks define MoEs: gates, experts, and a probabilistic weighting to combine expert predictions. Learning proceeds by finding a soft partitioning of the input space and assigning partitions to experts performing well on the partition. In this setting, the model response is then a sum of the experts' outputs, weighted by how confident the gate is in the expert's opinion. Yuksel et al.\ (2012) \cite{Yuksel2012} provide an extensive overview of recent developments in MoEs. The approach we propose allows learning such models, but also has applications to more general decision-making settings such as reinforcement learning. Ghosh et al.\ (2017) \cite{Ghosh2017} recently applied the divide-and-conquer principle to reinforcement learning. They argue that dividing a central policy into sub-policies improves the learning phase by improving sample efficiency. To evaluate this approach they assume \emph{pre-defined} partitions on the action and state space on which they train local policies. The information-theoretic constraints during training enforce similarity between the local policies such that a single central policy arises as weighted combination of all local policies. In contrast, in our approach all posterior expert policies remain close to their priors thereby minimizing their informational surprise. This mechanism leads to the emergence of specialized policies. In effect, this enforces the local policies to be as diverse as possible. Crucially, in our method the partitioning is not predefined but a result  of the optimization process itself.

Untangling the underlying structure in control systems usually requires a-priori knowledge of the system dynamics, e.g.,\ \cite{Abramova2012,Randlov2000,Yoshimoto2005}. The algorithm proposed by Abramova et al.\ (2012) \cite{Abramova2012} splits the state space of the inverted pendulum into predefined bins to fit a linear control policy to  stabilize individually. Then, the authors suggest to control the whole system by learning a selection policy over the given linear controllers. In contrast to this, our approach relies only on the reward signal to learn the selection policy and the linear control policy simultaneously. This fact alone poses a difficult learning problem as both system parts have to adjust to one another on different timescales. Other decentralized approaches (e.g.,\ Allamraju \& Chowdhary (2017) \cite{allamraju2017communication}) have trained separate decentralized models to fuse them into a single model. In contrast, our method learns sub-policies that act on their own.

Most other methods for meta-learning such as the work of \cite{Finn2017model} and \cite{ravi2017optimization}  find an initial parametrization of a single learner, such that the agent can adapt quickly to new problems. This initialization represents prior knowledge and can be regarded as an abstraction over related tasks and our method takes this idea one step further by finding a possibly disjunct set of such compressed task properties. Another way of thinking of such abstractions by lossy compression is to go from a task-specific posterior to a task-agnostic prior strategy. By having a set of priors the task specific information is available more locally then with a single prior, as in MAML \cite{Finn2017model} and the work of \cite{ravi2017optimization}. In principle, this can help to adapt within fewer iterations. Thus our method can be seen as the general case of such monolithic meta-learning algorithms. Instead of learning similarities within a problem, we can also try to learn similarities between different problems (e.g.,~ different classification datasets), as is described in the work of \cite{yao2019hierarchically}. In this way, the partitioning is governed by different tasks, where our study however focuses on discovering meta-information within the same task family, where the meta-partitioning is determined solely by the optimization process and can thus potentially discover unknown dynamics and relations within a task family.

\section{Conclusion}
In summary, we introduce and evaluate a promising novel on-line learning paradigm for hierarchical multi-agent systems. The main idea of our approach is an optimal soft partitioning by considering the agents' information constraints. The partitioning is automatic without relying on any prior information about the underlying problem structure or control dynamics in the case of model free learning. This makes our model abstract and principled. We apply it on a variety of tasks including multi-agent decision-making, mixture-of-expert regression, and divide-and-conquer reinforcement learning. We have extended this idea to a novel information-theoretic approach to meta-learning. In effect, the hierarchical structure equips the system with optimal initializations covering the input space which facilitates quick adaptation to new similar tasks. To build this hierarchical structure, we have proposed feature extraction models for classification, regression and reinforcement learning, that are able to extract task relevant information efficiently, invariant to the order of the inputs. The main strength of our approach is that it follows from simple principles that give rise to a large range of applications. Moreover, we can interpret the system performance by studying the information processing both at the selection stage and at the expert level, as we have shown by analyzing the discovered meta-knowledge. This can help to alleviate the problems inherent to black-box-approaches, for example based on deep neural networks.

\textbf{Contributions:} H.H.~ and D.A.B~ conceived the project, H.H.~ designed and implemented the algorithms and experiments and evaluated the results. Both authors discussed the results and wrote the manuscript. Both authors read and approved the manuscript.

\textbf{Funding:} This research was supported by the European Research Council, grant number ERC-StG-2015-ERC, Project
ID: 678082, ``BRISC: Bounded Rationality in Sensorimotor Coordination".

\textbf{Conflicts of Interest:} The authors declare no conflicts of interest.

\bibliographystyle{plain}
\bibliography{bibliography.bib}

\appendix
\section{Experimental Details}
\label{app:experiments}
\addcontentsline{toc}{section}{Appendices}
\renewcommand{\thesubsection}{\Alph{subsection}}
We implemented all experiments in Tensorflow and ran experiments working with images on a single NVIDIA Tesla V100-SXM2 32GB GPU\@. We implemented our experiments in TensorFlow \cite{abadi2016tensorflow}, Scikit-learn \cite{scikit-learn} and OpenAI Gym \cite{Brockman2016}.

\subsection{Non-Meta-Learning Setting}
For classification (Section \ref{sec:experimentsclass}) we use a linear predictor as experts, such that the expert's response is a linear combination of the inputs weighted by some learned parameters $\omega$: $y = \omega^Tx$ and the selector is a two-layer MLP with 10 units each and tanh non-linearities. 

In reinforcement learning (Section \ref{sec:experimentsrl}) the selection network is a two layer network with 32 units per layer and tanh non-linearities. Experts are two layer networks with tanh non-linearities that learn log-variance and mean of a Gaussian distribution to predict the control signal. The critic networks have the same architecture but learn directly the value function. As the action space is continuous in the interval [-1,1] we learn $\mu$ and $\log(\sigma)$ of a Gaussian by parameterizing the distribution with a neural network. We sample actions by re-parameterizing the distribution to $p(a) = \mu + \sigma \epsilon$, where $\epsilon \thicksim \altmathcal{N}(0,1)$, so that the distribution is differentiable w.r.t. the network outputs (``re-parametrization trick`` introduced by Kingma and Welling (2013) \cite{Kingma2013}).

We train all networks using Adam \cite{kingma2014adam} with a learning rate of $3\cdot10^{-4}$ with Mini-Batch sizes of 32 and sample 1024 data from each task (``Half Moon" etc.) for 10000 episodes. We average the results presented are over 10 random seeds. 

\subsection{Meta-Learning Setting}
\label{app:meta}

\begin{figure*}[t!]
\centering
\includegraphics[width=0.95\textwidth, trim={0cm 21cm 3.25cm 0cm}, clip]{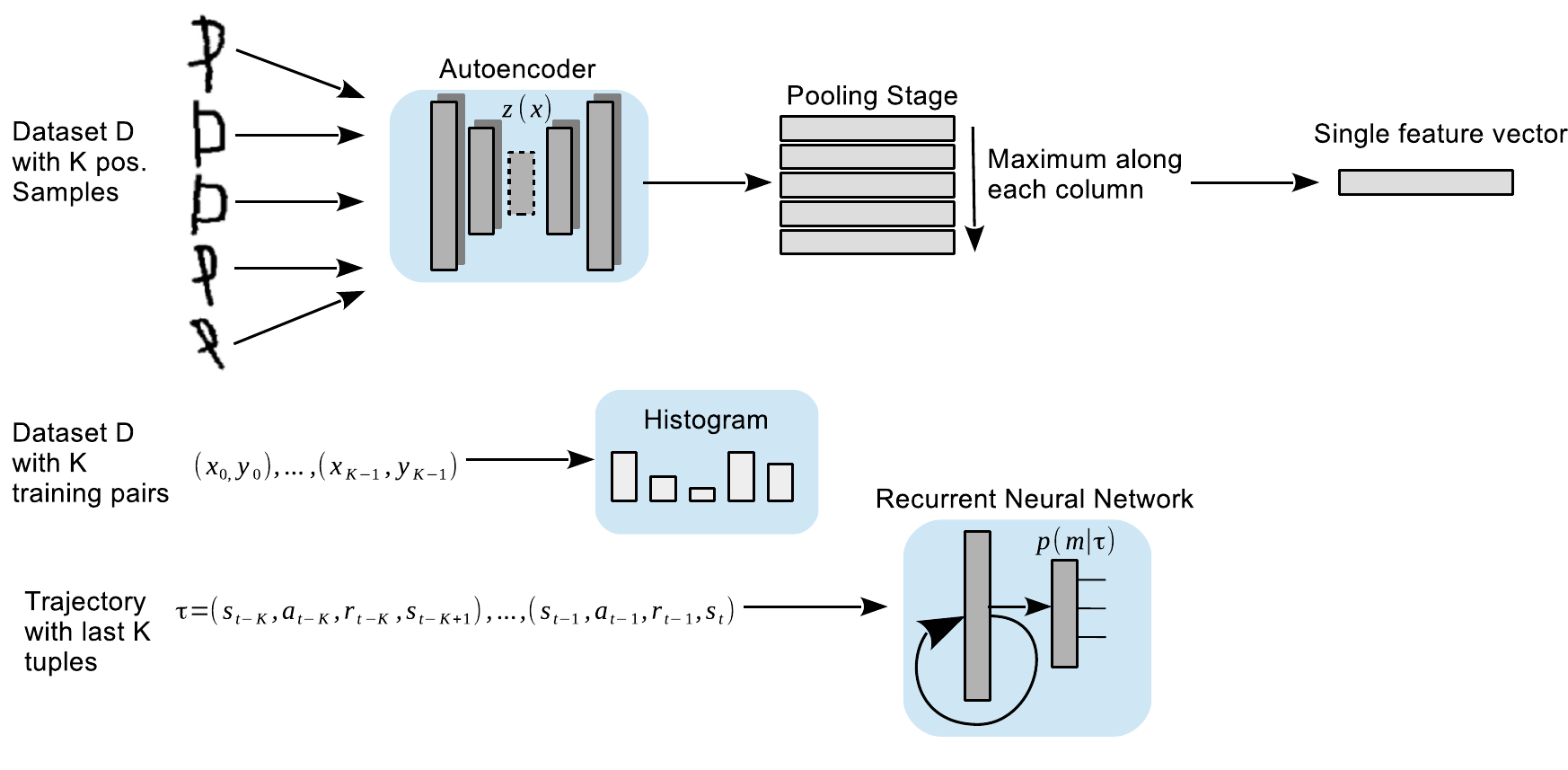}
\caption{Dataset features in the meta-learning setup. In supervised meta-learning we pass $K$ samples through an convolutional autoencoder producing a $K \times N$ feature matrix, where $N$ is the innermost feature dimension of the autoencoder's bottleneck layer. To find a single feature vector, we apply a max operator along the columns of the feature matrix. For regression we compute a histogram over the data points. In reinforcement learning, at the start of a trial we sample an expert according to the prior $p(m)$, run the policy for $K$ time steps and use these $K$ $(s,a,r,s')$ tuples as input to an recurrent neural network, that acts the selection network to find an expert, which remains active until the environment resets.}
\label{fig:features}
\end{figure*}

The features that we use in the meta-learning setting for the selector in the different learning scenarios are depicted in Figure~\ref{fig:features}.
For regression (Section \ref{sec:experimentssine}) we use a two layer selection network with 16 units each followed by tanh non-linearities. The experts are shallow neural networks with a single hidden layer that learn log-variance and mean of a Gaussian distribution which they use for prediction. We use the ``Huber Loss" instead of MSE as it is more robust \cite{balasundaram2019robust}. We optimize all networks using Adam \cite{kingma2014adam}. We set $\beta_1 = 25$ and $\beta_2 = 1.25$.

For Omniglot (Section \ref{sec:experimentsfew}) we followed the design of Vinyals et al.\, \cite{vinyals2016matching} but reduce the number of blocks to one. We used a single convolutional block consisting of 32 3$\times$3 filters with strided convolutions followed by a batch normalization layer and a ReLu non-linearity. The output is fed into a softmax layer giving a distribution over classes. During training we used a meta-batch size of 16. The convolutional autoencoder is a 3 layer network consisting of 16, 16, and 4 filters each with size 3$\times$3 with strided convolutions followed by a leaky ReLu non-linearity. The layers are mirrored by de-convolotional layers to reconstruct the image. This results in an image embedding with dimensionality 64. The selection network is a two layer network with 32 units, followed by a ReLu non-linearity, a dropout layer \cite{srivastava2014dropout} per layer and is fed into a softmax normalization to produce a distribution over the experts. For 8 and more experts we add a third layer with 32 units. To improve generalization we add a MaxNorm regularization on the weights. We augment the dataset by rotating each image in 90, 180, and 270 degrees resulting in 80 images per class. We also normalize the images two be in (0,1) range. We evaluate our method by resetting the system to the state after training and allow for 10 gradient updates and report the final accuracy. We train all networks using Adam \cite{kingma2014adam} with a learning rate of $3 \cdot 10^{-4}$. In this experiment we set $\beta_1 = 20.0$ and $\beta_2 = 2.5$ for 2 and 4 experts and $\beta_1 = 50.0$ and $\beta_2 = 1.25$ for 8 and 16 experts.

In meta reinforcement learning (Section \ref{sec:experimentsmetarl}) the selector's actor and critic net are build of RNNs with 200 hidden units each. The critic  is trained to minimize the Huber loss between the prediction and the cumulative reward. The experts are two layer networks with 64 units each followed by ReLu non-linearities and used to learn the parameters of a Gaussian distribution. The critics have the same architecture (except for the output dimensionality). The actors learning rate is set to $10^{-4}$ and the critics to $10^{-3}$. We optimize all networks using Adam \cite{kingma2014adam}.  We set $\beta_1 = 25.0$ and $\beta_2 = 2.5$.

To evaluate MAML on 2-way $N$-shot omniglot dataset we used a inner learning rate of $\alpha = 0.05$ and one inner update step per iteration for all settings. We used a single convolutional block followed by a fully connected layer with 64 units and a ReLU non-linearity. For matching networks we used the same architecture. Note, that we reduce the number of layers to make the tests comparable. Using the suggested architectures \cite{Finn2017model,vinyals2016matching} we achieve classification accuracy $\geq 95\%$.

\section{DKL between two Normal Wishart Distributions}
\label{app:wishartdkl}
KL divergence from $Q$ to $P$ is defined as $\DKL(P \Vert Q) = \int p(m) \log \frac{p(m)}{q(m)}dx$ and a Normal-Wishart distribution is defined as
\begin{equation*}
f(\bm{\mu},\bm{\Lambda} \vert \bm{\omega}, \lambda, \bm{W}, \nu) = \altmathcal{N} \left(\bm{\mu} \vert \bm{\omega}, (\lambda \bm{\Lambda})^{-1} \right) \altmathcal{W}(\bm{\Lambda} \vert \bm{W}, \nu),
\end{equation*}
where $\bm{\omega} \in \mathbb{R}^D, \bm{W} \in \mathbb{R}^{D\times D},\nu > D - 1, \lambda > 0$ are the parameters of the distribution. We optimize over $\bm{\omega}$ and $\bm{W}$, which makes part of the $\DKL$ terms constant. So we have:
\begin{eqnarray*}
D_{\mathrm{KL}} \left[ \altmathcal{N}_{0}(\bm{\mu}) \altmathcal{W}_{0}(\bm{\Lambda}) \Vert \altmathcal{N}_{1}(\bm{\mu}) \altmathcal{W}_{1}(\bm{\Lambda}) \right] &=& \int \altmathcal{N}_{0}(\bm{\mu}) \altmathcal{W}_{0}(\bm{\Lambda}) \log \frac{\altmathcal{N}_{0}(\bm{\mu}) \altmathcal{W}_{0}(\bm{\Lambda})}{\altmathcal{N}_{1}(\bm{\mu}) \altmathcal{W}_{1}(\bm{\Lambda})}d\bm{\mu}d\bm{\Lambda}
\end{eqnarray*}
Now let  
\begin{eqnarray*}
p(\bm{\mu}, \bm{\Lambda}) &=& p(\bm{\mu}\vert\bm{\Lambda})p(\bm{\Lambda}) = \altmathcal{N}\left( \bm{\mu} \vert \bm{\mu}_{0}, (\lambda_{0} \bm{\Lambda})^{-1} \right) \altmathcal{W} \left( \bm{\Lambda} \vert \bm{W}_{0}, \nu_{0} \right)\\
q(\bm{\mu}, \bm{\Lambda}) &=& q(\bm{\mu}\vert\bm{\Lambda})q(\bm{\Lambda}) = \altmathcal{N}\left( \bm{\mu} \vert \bm{\mu}_{1}, (\lambda_{1} \bm{\Lambda})^{-1} \right) \altmathcal{W} \left( \bm{\Lambda} \vert \bm{W}_{1}, \nu_{1} \right).
\end{eqnarray*}
We can find the $\DKL$ as follows:
\begin{eqnarray*}
D_{\mathrm{KL}} \left[ p(\bm{\mu}, \bm{\Lambda}) \Vert q(\bm{\mu}, \bm{\Lambda}) \right] 
&=& \int_{\bm{\mu}} \int_{\bm{\Lambda}} p(\bm{\mu}, \bm{\Lambda}) \log \frac{p(\bm{\mu}, \bm{\Lambda})}{q(\bm{\mu}, \bm{\Lambda})} d\bm{\mu} d\bm{\Lambda} \\
&=& \int_{\bm{\mu}} \int_{\bm{\Lambda}} p(\bm{\mu} \vert \bm{\Lambda}) p(\bm{\Lambda}) \log \frac{p(\bm{\mu} \vert \bm{\Lambda}) p(\bm{\Lambda})}{q(\bm{\mu} \vert \bm{\Lambda}) q(\bm{\Lambda})} d\bm{\mu} d\bm{\Lambda} \\
&\stackrel{\text{indep.}}{=}& \int_{\bm{\mu}} \int_{\bm{\Lambda}} p(\bm{\mu} \vert \bm{\Lambda}) p(\bm{\Lambda}) \log \frac{p(\bm{\mu} \vert \bm{\Lambda})}{q(\bm{\mu} \vert \bm{\Lambda})} d\bm{\mu} d\bm{\Lambda} \\ & & + \int_{\bm{\mu}} \int_{\bm{\Lambda}} p(\bm{\mu} \vert \bm{\Lambda}) p(\bm{\Lambda}) \log \frac{p(\bm{\Lambda})}{q(\bm{\Lambda})} d\bm{\mu} d\bm{\Lambda}\\
&=& \int_{\bm{\Lambda}} p(\bm{\Lambda}) \left[\int_{\bm{\mu}} p(\bm{\mu} \vert \bm{\Lambda}) \log \frac{p(\bm{\mu} \vert \bm{\Lambda})}{q(\bm{\mu} \vert \bm{\Lambda})} d\bm{\mu} \right] d\bm{\Lambda} \\ & & + \int_{\bm{\Lambda}} p(\bm{\Lambda}) \log \frac{p(\bm{\Lambda})}{q(\bm{\Lambda})} d\bm{\Lambda}\\
&\stackrel{\text{def.}}{=}& \mathbb{E}_{p(\bm{\Lambda})} \left[ \DKL \left[ p(\bm{\mu} \vert \bm{\Lambda}) \Vert q(\bm{\mu} \vert \bm{\Lambda}) \right] \right] + \DKL \left[ p(\bm{\Lambda}) \Vert q(\bm{\Lambda}) \right].\\
\end{eqnarray*}
where $\mathbb{E}_{p(\bm{\Lambda})}[X] = \nu\bm{W}$ and $\DKL$ between two Normal Distributions $p(\bm{\mu} \vert \bm{\Lambda})$ and $q(\bm{\mu} \vert \bm{\Lambda})$ is
\begin{eqnarray*}
\DKL \left[ p(\bm{\mu} \vert \bm{\Lambda}) \Vert q(\bm{\mu} \vert \bm{\Lambda}) \right]
&=& \frac{1}{2} \bigg[ \mathrm{tr}\left((\lambda_q \bm{\Lambda})(\lambda_p \bm{\Lambda})^{-1} \right) + \left( \bm{\mu}_{q} - \bm{\mu}_{p} \right)^{\top}(\lambda_q \bm{\Lambda}) \left( \bm{\mu}_{q} - \bm{\mu}_{p} \right) \\ & & - D + \log \frac{\Vert\lambda_p \bm{\Lambda}\Vert}{\Vert\lambda_q \bm{\Lambda}\Vert}\bigg] \\
&=& \frac{1}{2} \left[ D \frac{\lambda_q}{\lambda_p} + \left( \bm{\mu}_{q} - \bm{\mu}_{p} \right)^{\top} \lambda_q \bm{\Lambda} \left( \bm{\mu}_{q} - \bm{\mu}_{p} \right) - D + D \log \frac{\lambda_q}{\lambda_p} \right],
\end{eqnarray*}
so we have
\begin{eqnarray*}
\mathbb{E}_{p(\bm{\Lambda})} \left[ D_{\mathrm{KL}} \left[ p(\bm{\mu} \vert \bm{\Lambda}) \Vert q(\bm{\mu} \vert \bm{\Lambda}) \right] \right] 
&=& \frac{\lambda_{q}}{2} \left( \bm{\mu}_{q} - \bm{\mu}_{p} \right)^{\top} \nu_{p} \mathbf{W}_{p} \left( \bm{\mu}_{q} - \bm{\mu}_{p} \right) \\ & & + \underbrace{\frac{D}{2} \left( \frac{\lambda_{q}}{\lambda_{p}} - \log \frac{\lambda_{q}}{\lambda_{p}} - 1 \right)}_{\text{constant term}}.
\end{eqnarray*}
The $\DKL$ between two Wishart Distributions $p(\bm{\Lambda})$ and $q(\bm{\Lambda})$ is
\begin{eqnarray*}
D_{KL}[p(\bm{\Lambda}) \\vert  q(\bm{\Lambda})] &=& H(p(\bm{\Lambda}), q(\bm{\Lambda})) - H(p(\bm{\Lambda})) \\[6pt]
 &=& -\frac{\nu_q} 2 \log \vert \bm{W}_q^{-1} \bm{W}_p\vert   + \frac{\nu_p}{2}(\text{tr}(\bm{W}_q^{-1} \bm{W}_p) - D) \\ & & + \underbrace{\log \frac{\Gamma_D\left(\frac{\nu_q} 2 \right)}{\Gamma_D\left(\frac{\nu_p} 2 \right)} + \tfrac{\nu_p - \nu_q } 2 \psi_D\left(\frac{\nu_p} 2\right)}_{\text{constant term}}
\end{eqnarray*}
where $\Gamma_D(m)$ is the multivariate Gamma distribution and $ \psi_D(m)$ is the derivative of the log of the multivariate Gamma distribution, each parameterized by the dimensionality of $\bm{\mu}$ and $\bm{W}$ denoted by $D$. As we keep $\nu$ and $\lambda$ fixed, we can use an estimate, which is only off by constant factor $C$:
\begin{equation}
\begin{split}
\DKL \left[ p(\bm{\mu}, \bm{\Lambda}) \Vert q(\bm{\mu}, \bm{\Lambda}) \right] = \frac{\lambda_{q}}{2} \left( \bm{\mu}_{q} - \bm{\mu}_{p} \right)^{\top} \nu_{p} \mathbf{W}_{p} \left( \bm{\mu}_{q} - \bm{\mu}_{p} \right) - \\ \frac{\nu_q} 2 \log \vert \bm{W}_q^{-1} \bm{W}_p\vert   + \frac{\nu_p}{2}(\text{tr}(\bm{W}_q^{-1} \bm{W}_p) - D) + C
\end{split}
\end{equation}
where
$C = \frac{D}{2} \left( \frac{\lambda_{q}}{\lambda_{p}} - \log \frac{\lambda_{q}}{\lambda_{p}} - 1 \right) + \log \frac{\Gamma_D\left(\frac{\nu_q} 2 \right)}{\Gamma_D\left(\frac{\nu_p} 2 \right)} + \tfrac{\nu_p - \nu_q } 2 \psi_D\left(\frac{\nu_p} 2\right)$

\section{Additional Results}
\label{app:results}

\begin{figure*}[h]
\label{fig:sinepartition16}
\centering
\includegraphics[width=0.75\textwidth]{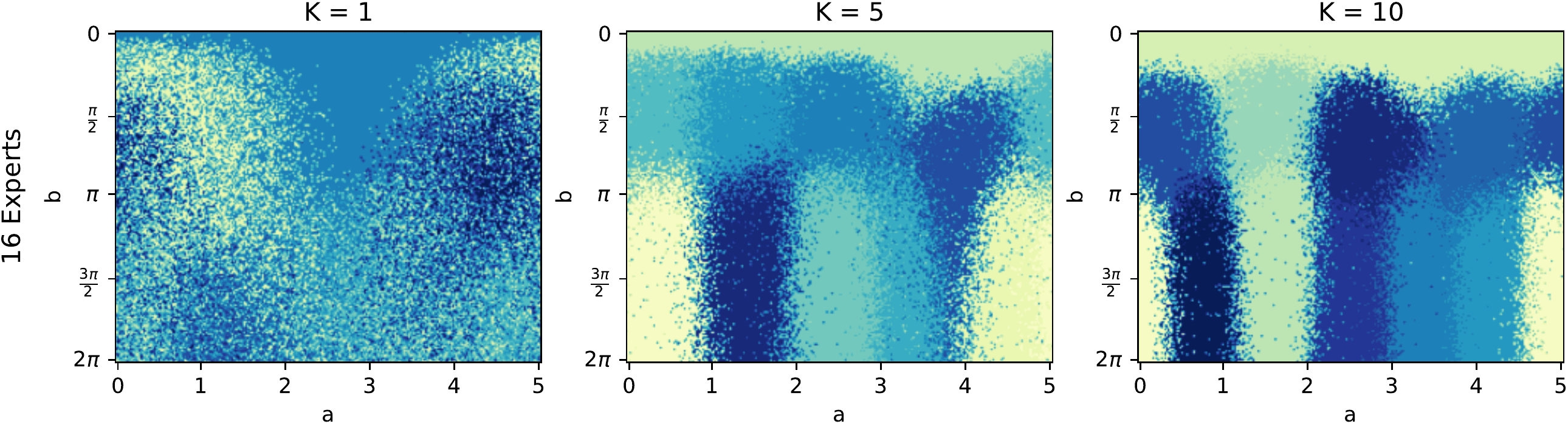}
\caption{Here we show the soft-partition found by the selection policy for the sine prediction problem $y = a\cdot\sin(x + b)$ for 16 experts, where we sample $a,b$ uniformly at each trial and each color represents an expert.}
\end{figure*}
\begin{figure*}[b]
  \begin{minipage}{.245\textwidth}
  \centering
    \includegraphics[width=0.95\linewidth]{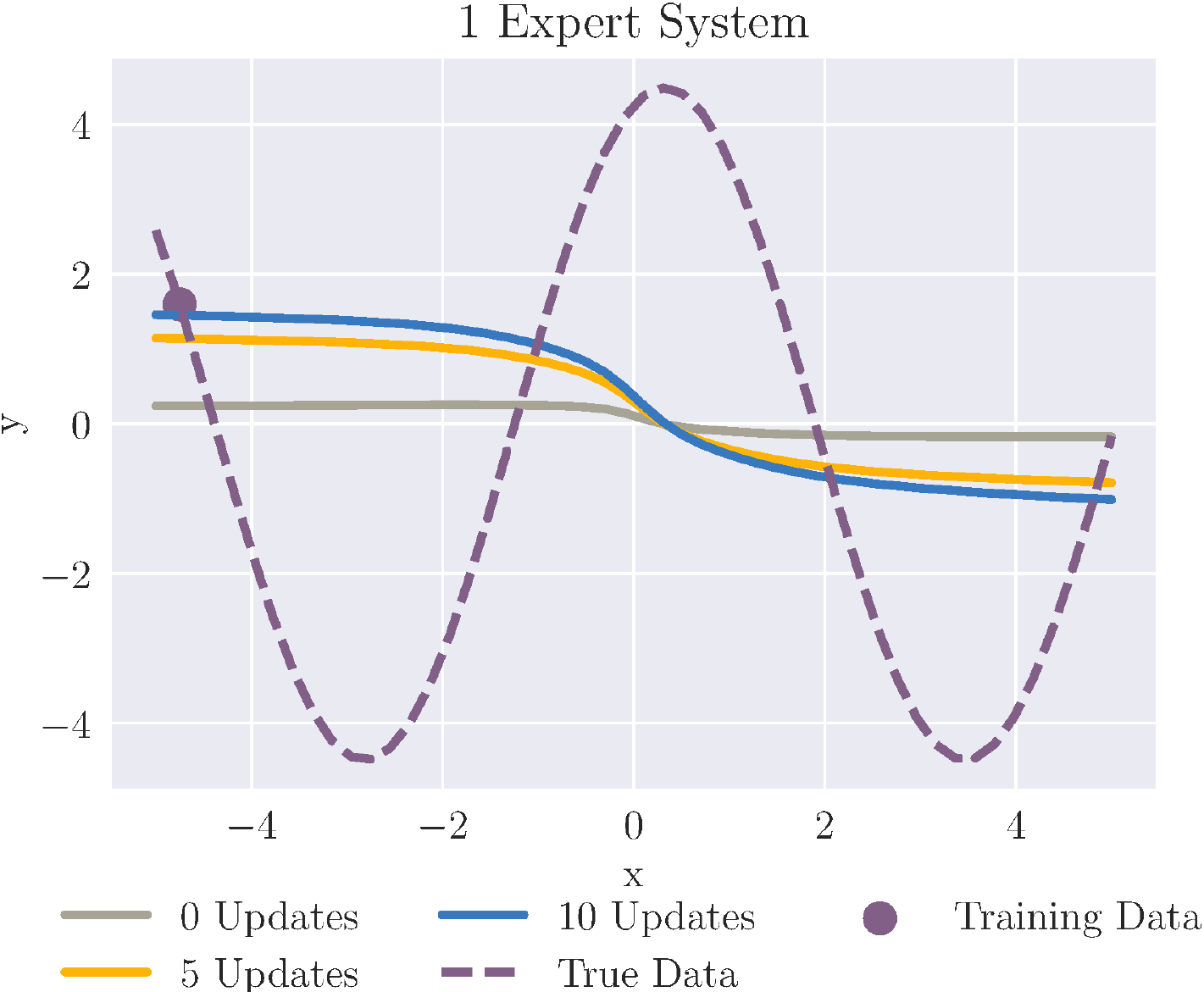}
  \end{minipage}
  \begin{minipage}{.245\textwidth}
  \centering
    \includegraphics[width=0.95\linewidth]{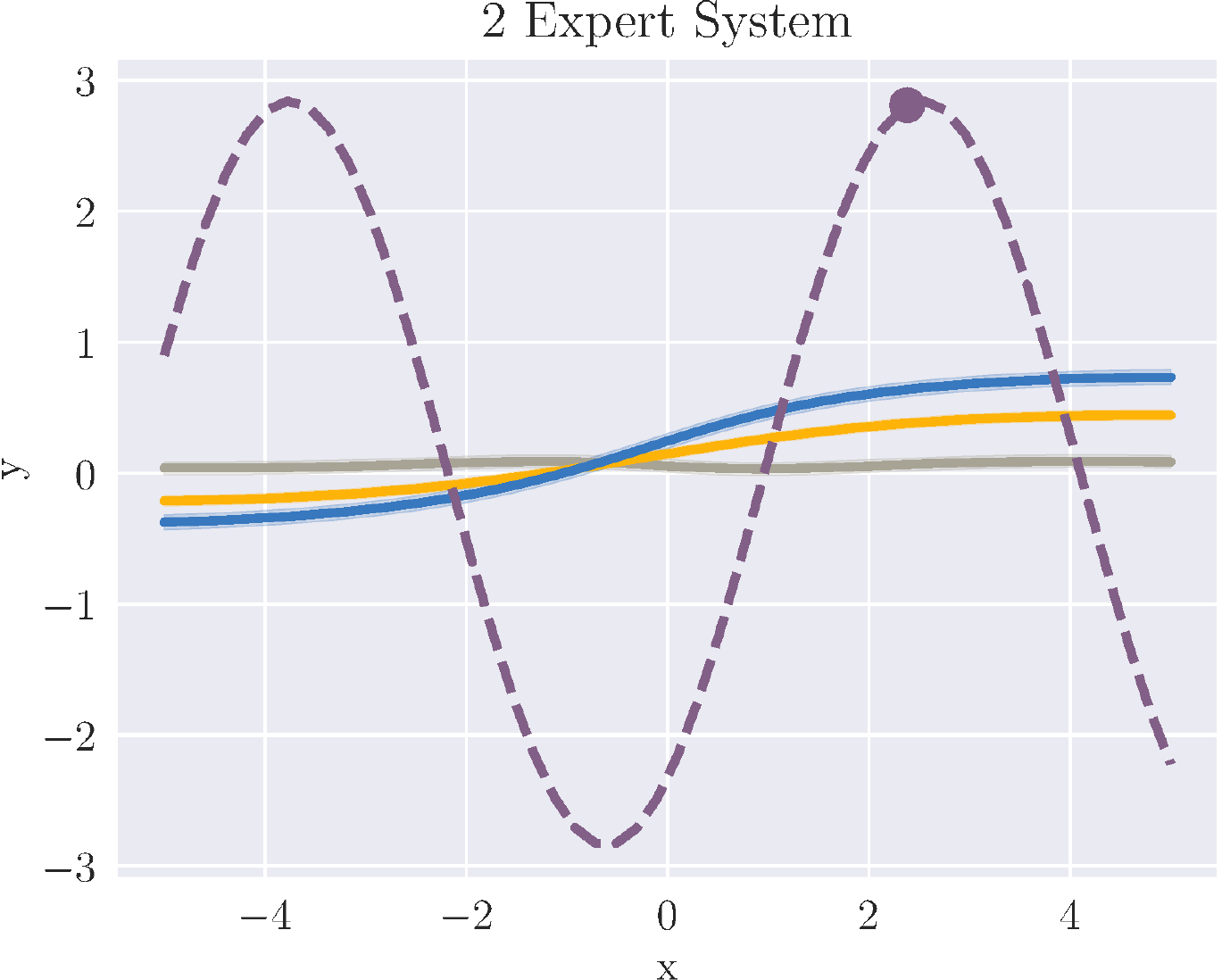}
  \end{minipage}
    \begin{minipage}{.245\textwidth}
    \centering
    \includegraphics[width=0.95\linewidth]{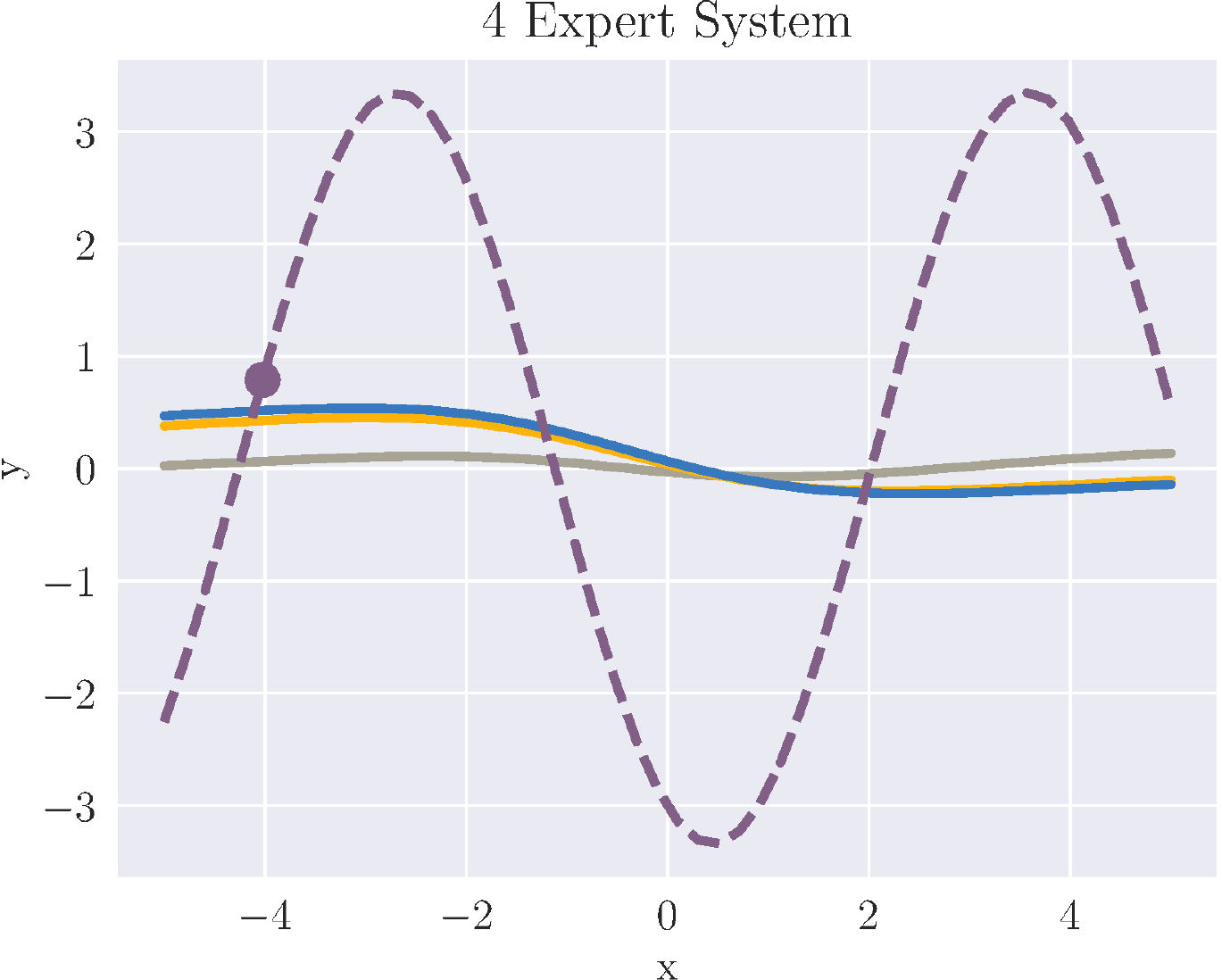}
  \end{minipage}
    \begin{minipage}{.245\textwidth}
    \centering
    \includegraphics[width=0.95\linewidth]{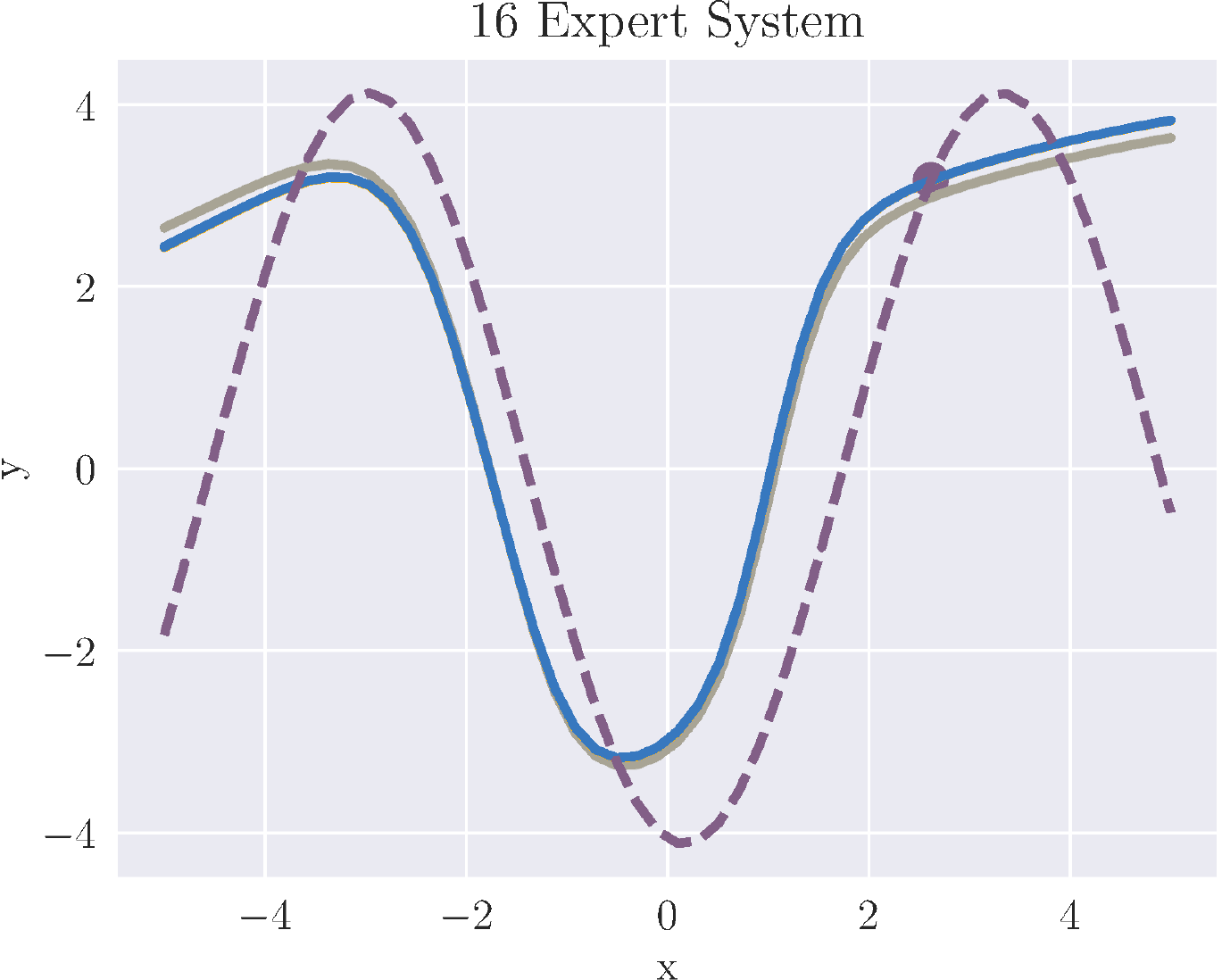}
  \end{minipage}
    \begin{minipage}{.245\textwidth}
    \centering
    \includegraphics[width=0.95\linewidth]{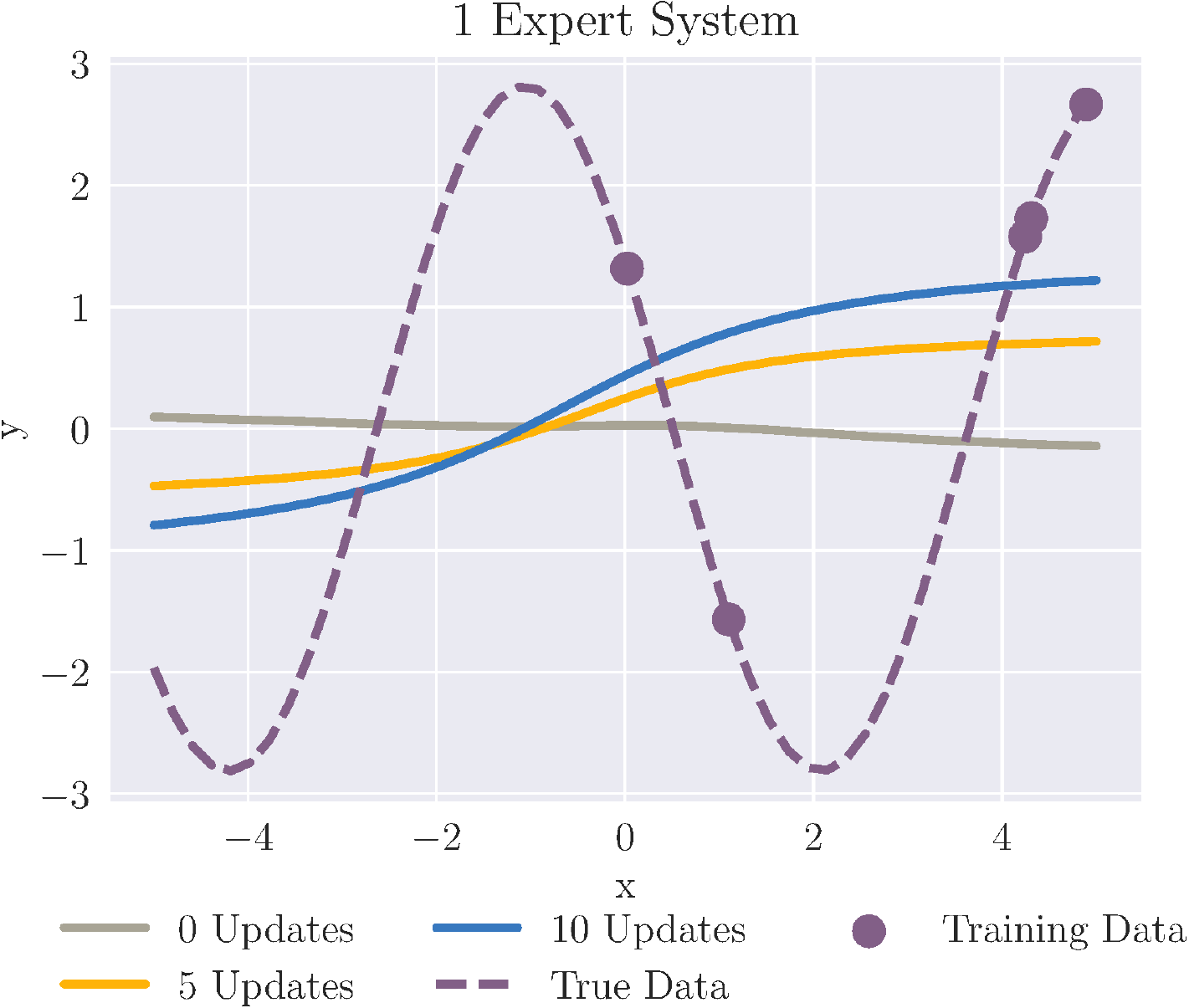}
  \end{minipage}
  \begin{minipage}{.245\textwidth}
  \centering
    \includegraphics[width=0.95\linewidth]{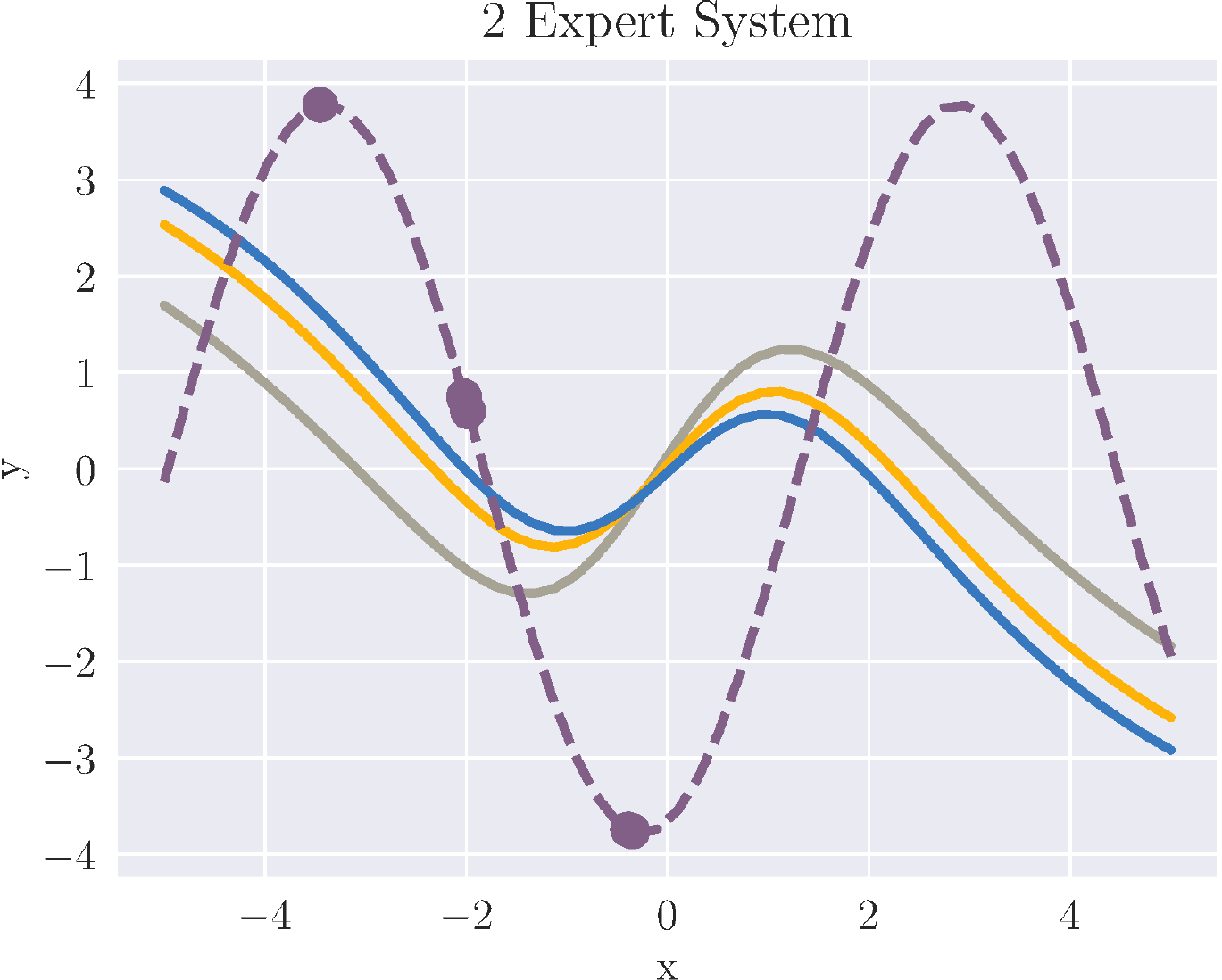}
  \end{minipage}
    \begin{minipage}{.245\textwidth}
    \centering
    \includegraphics[width=0.95\linewidth]{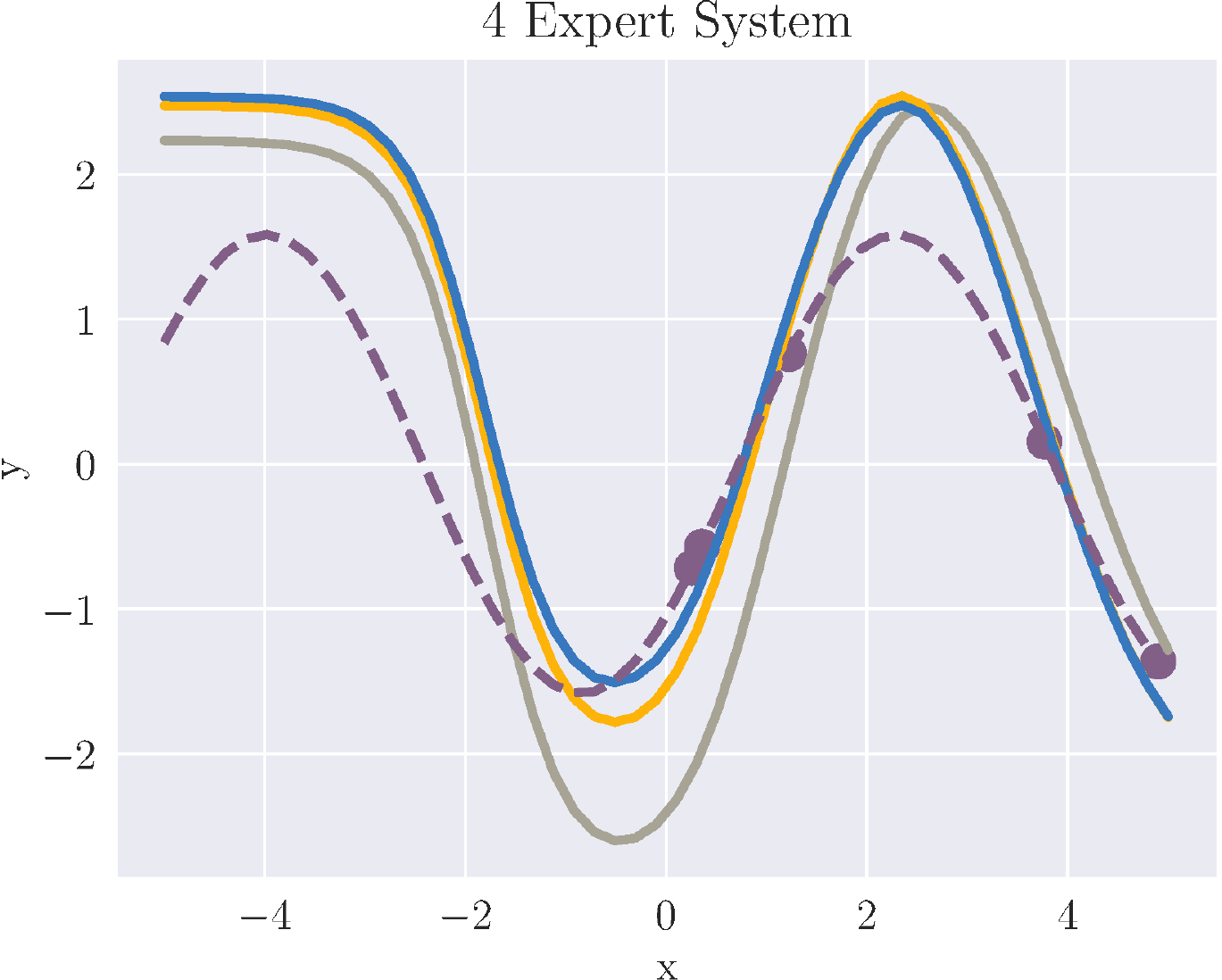}
  \end{minipage}
    \begin{minipage}{.245\textwidth}
    \centering
    \includegraphics[width=0.95\linewidth]{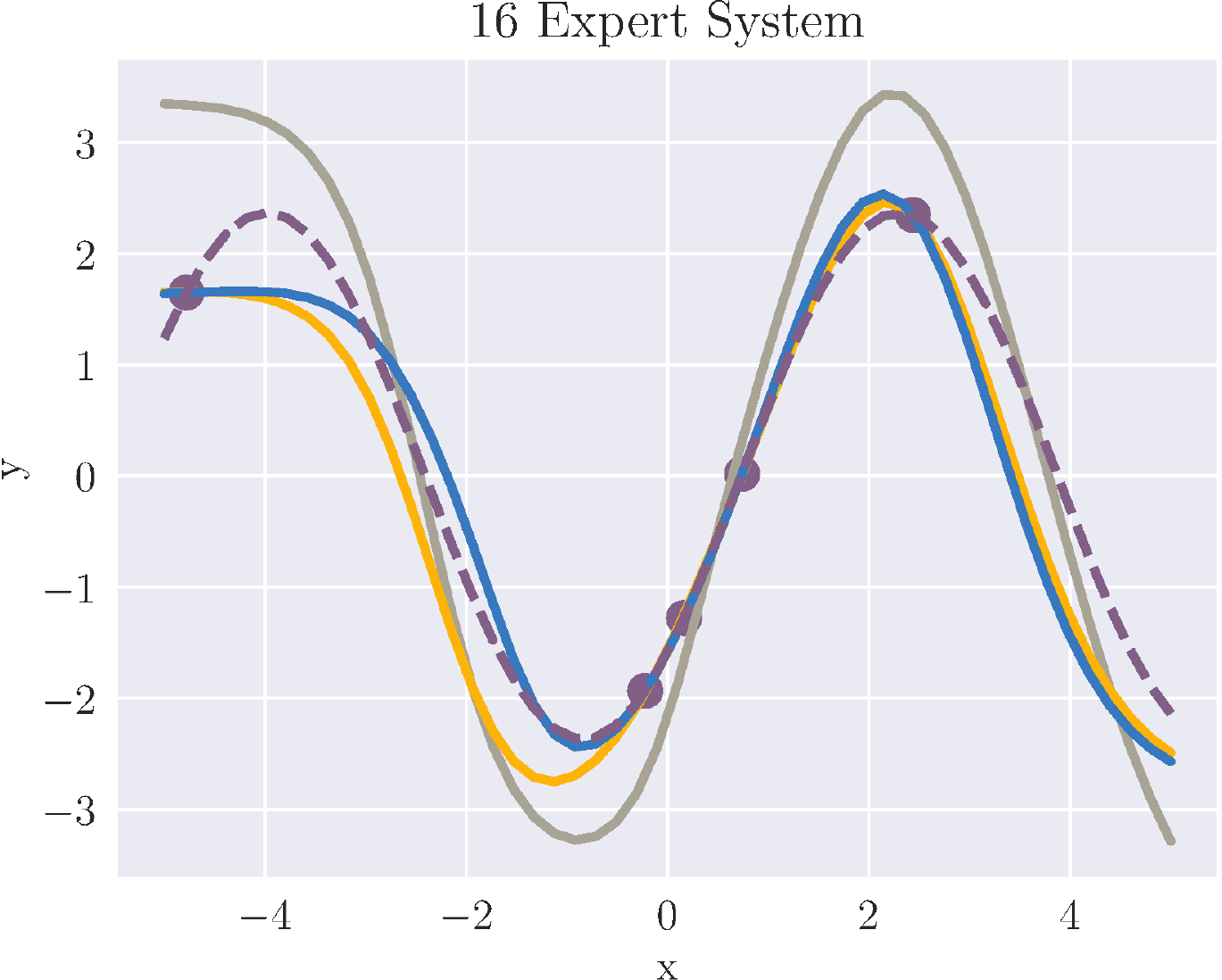}
  \end{minipage}
\caption{Here show how the system is able to adapt to new problems as the number of experts increases additional results for the $K = 1$ (upper row) and $K = 5$ (lower row) settings.} 
\label{fig:sineapp}
\end{figure*}

\end{document}